\documentclass[letterpaper]{article} % DO NOT CHANGE THIS
\usepackage{aaai23}  % DO NOT CHANGE THIS
\usepackage{times}  % DO NOT CHANGE THIS
\usepackage{helvet}  % DO NOT CHANGE THIS
\usepackage{courier}  % DO NOT CHANGE THIS
\usepackage[hyphens]{url}  % DO NOT CHANGE THIS
\usepackage{graphicx} % DO NOT CHANGE THIS
\usepackage{natbib}  % DO NOT CHANGE THIS AND DO NOT ADD ANY OPTIONS TO IT
\usepackage{caption} % DO NOT CHANGE THIS AND DO NOT ADD ANY OPTIONS TO IT
\usepackage{algorithm}
\usepackage{algorithmic}

\usepackage{amsmath,amssymb}
\usepackage{caption}
\usepackage{subcaption}
\usepackage{epsfig}
\usepackage{url}
\usepackage{comment}
\usepackage{color}
\usepackage{enumitem}
\usepackage{booktabs}

\usepackage{graphicx}
\usepackage{bm}
\usepackage{multirow}
\usepackage{wrapfig}
\usepackage{xcolor}
\usepackage{color, colortbl}

\usepackage{newfloat}
\usepackage{listings}
\DeclareCaptionStyle{ruled}{labelfont=normalfont,labelsep=colon,strut=off} % DO NOT CHANGE THIS
\floatstyle{ruled}
\newfloat{listing}{tb}{lst}{}
\floatname{listing}{Listing}
\title{Deep Parametric 3D Filters for Joint Video Denoising and Illumination Enhancement in Video Super Resolution}
\author{
Xiaogang Xu\textsuperscript{\rm 1,2} \quad
Ruixing Wang \textsuperscript{\rm 2} \quad
Chi-Wing Fu \textsuperscript{\rm 1} \quad
Jiaya Jia \textsuperscript{\rm 1,2} 
}
\affiliations{
\textsuperscript{\rm 1} The Chinese University of Hong Kong \quad \textsuperscript{\rm 2} SmartMore\\
\{xgxu, cwfu, leojia\}@cse.cuhk.edu.hk \quad ruixing.wang@smartmore.com
}

\usepackage{bibentry}
\begin{document}

\maketitle

\begin{abstract}
	Despite the quality improvement brought by the recent methods, video super-resolution (SR) is still very challenging, especially for videos that are low-light and noisy.
	The current best solution is to subsequently employ best models of video SR, denoising, and illumination enhancement, but doing so often lowers the image quality, due to the inconsistency between the models.
	This paper presents a new parametric representation called the {\em Deep Parametric 3D Filters\/} (DP3DF), which incorporates local spatiotemporal information to enable simultaneous denoising, illumination enhancement, and SR efficiently in a single encoder-and-decoder network.
	Also, a dynamic residual frame is jointly learned with the DP3DF via a shared backbone to further boost the SR quality.
	We performed extensive experiments, including a large-scale user study, to show our method's effectiveness.
	Our method {\em consistently\/} surpasses the best state-of-the-art methods on all the challenging real datasets with top PSNR and user ratings, yet having a very fast run time. The code is available at \url{https://github.com/xiaogang00/DP3DF}.
\end{abstract}

\section{Introduction}

The goal of video super resolution (SR) is to produce high-resolution videos from low-resolution video inputs.
While promising results are demonstrated on general videos, existing approaches typically do not work well on videos that are low-light and noisy. Yet, such a setting is very common in practice,~e.g., applying SR to enhance noisy videos taken in a dark and high-contrast environment.

Fundamentally, video denoising and video illumination enhancement are very different tasks from video SR: the former deals with noise and brightness in videos whereas the latter deals with the video resolution.
Hence, to map a low-resolution, low-light, and noisy (LLN) video to a high-resolution, normal-light, and noise-free (HNN) video, the current best solution is to collectively use the best network model of each task by cascading models in a certain order.

\begin{figure}[t]
	\centering
	\includegraphics[width=1.0\linewidth]{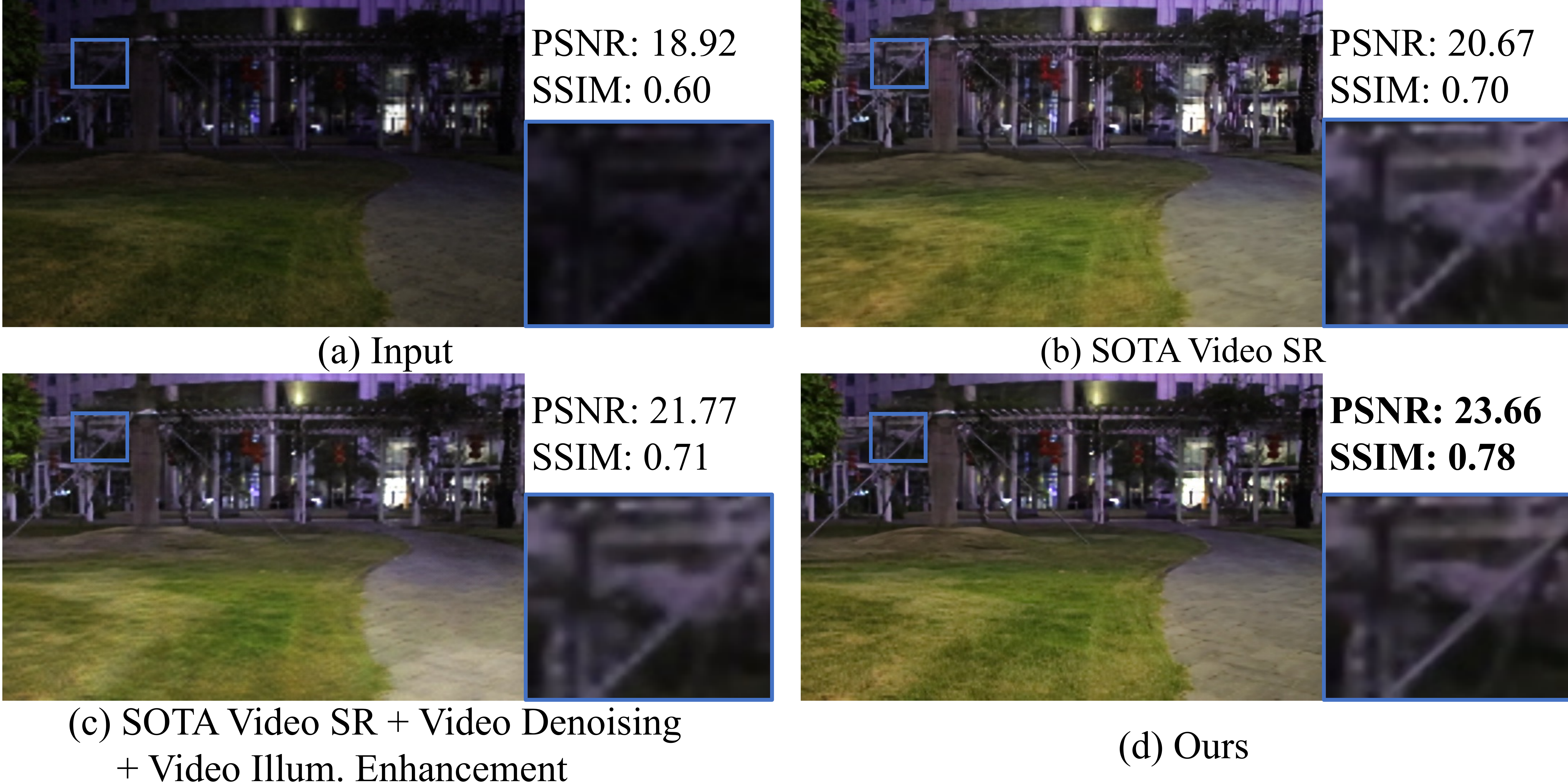}
	\vspace{-0.25in}
	\caption{An example frame (a) from a challenging underexposed video enhanced by 
		(b) a SOTA SR method~\cite{xiang2020zooming};
		(c) SOTA methods in video denoising~\cite{tassano2020fastdvdnet} $+$ video illumination enhancement~\cite{zhang2021learning} $+$ video SR~\cite{xiang2020zooming}; and
		(d) our approach.
		From the blown-up views, we can see that (d) is more sharp 
		with distinct contrast, less noise, and less aliasing vs. (b) \& (c).
		Please zoom to view the details.
	}
	\vspace{-0.2in}
	\label{fig:teaser}
\end{figure}

However, doing so has several drawbacks.
First, the network complexity is threefold, resulting in a slow inference, as we have to subsequently run three separate network models for denoising, illumination enhancement, and SR.
Also, as the three networks are trained separately, we cannot ensure their consistency,~e.g., artifacts from a preceding denoising or illumination enhancement network could be amplified by the subsequent SR network; see Fig.~\ref{fig:teaser} (c).
Alternatively, one may try to cascade and train all networks end-to-end.
Yet, existing networks for video SR, denoising, and illumination enhancement often take multiple frames as input and output only a single frame, so a subsequent network cannot obtain sufficient inputs from the preceding one.
Also, these networks are rather complex, so it is hard to fine-tune them together for high performance.

Another approach is to use parallel branches of different purposes in a framework. However, as the branches are separated from one another, their connections are weak for joint learning.
Also, the input to all branches should be identical, while the sizes of their outputs are inconsistent: the output size of the SR branch is larger than its input size, while the other branches have same input and output sizes.
Hence, how to achieve various purposes with one common branch and representation is worth to be considered.
Further, we eventually will need to infer the different branches to produce the final results, which is time costly.

In this paper, we present a new solution to map LLN videos to HNN videos within a single end-to-end network.
The core of our solution is the {\em Deep Parametric 3D Filter\/} ({\em DP3DF}), a novel dynamic-filter representation we formulated collectively for video SR, illumination enhancement, and denoising.
This is the first work that we are aware of in exploring an efficient architecture for simultaneous video SR, denoising, and illumination enhancement.
Beyond the existing works with dynamic filters, our DP3DF considers the burst from adjacent frames.
Hence, DP3DF can effectively exploit local spatiotemporal neighboring information and complete the mapping from LLN video to HNN video in a single encoder-and-decoder network.
Also, we show that general dynamic filters in existing works are just special cases of our DP3DF.
%; see Section~\ref{connection}.
Further, we set up an additional branch for learning dynamic residual frames on top of the core encoder-and-decoder network, so we can share the backbone for learning the DP3DF and residual frames to promote the overall performance.

To demonstrate the quality of our method, we conducted comprehensive experiments to compare our method with a rich set of state-of-the-art methods on two public video datasets SMID~\cite{chen2019seeing} and SDSD~\cite{sdsd}, which provide static and dynamic low- and normal-light video pairs.
Through various quantitative and qualitative evaluations, including a large-scale user study with 80 participants, we show the effectiveness of our DP3DF framework over SOTA SR methods and also different combinations of SOTA video methods on illumination enhancement, denoising, and SR, both quantitatively and qualitatively.
Our DP3DF framework surpasses the SOTA methods with top PSNR and user ratings consistently.
In summary, our contributions are threefold:
\begin{itemize}
	\item This is the first exploration of directly mapping LLN to HNN videos within a single-stage end-to-end network.
	\item This is the first work we are aware of that simultaneously achieves video SR, denoising, and illumination enhancement via our DP3DF representation.
	\item Extensive experiments are conducted on two real-world video datasets,demonstrating our superior performance.   %demonstrating the superior performance of our method.  
\end{itemize}

\vspace{-0.1in}
\section{Related work}
\noindent\textbf{Video SR.} \
Video SR aims to reconstruct a high-resolution frame from a low-resolution frame together with the associated adjacent frames. 
The key problem is on how to align the adjacent frames temporally with the center one.
Several video SR methods~\cite{caballero2017real,tao2017detail,sajjadi2018frame,wang2018learning,xue2019video} use optical flow for an explicit temporal alignment.
However, it is hard to obtain accurate flow and the flow warping may introduce artifacts in the aligned frames.
To leverage the temporal information, recurrent neural networks are adopted in some video SR methods~\cite{huang2017video,lim2017deep}, e.g., the convolutional LSTMs~\cite{shi2015convolutional}.
However, without an explicit temporal alignment, these RNN-based networks have limited capability in handling complex motions.
Later, dynamic filters and deformable convolutions are exploited for temporal alignment.
DUF~\cite{jo2018deep} utilizes a dynamic filter to implement simple temporal alignment without motion estimation, whereas TDAN~\cite{tian2020tdan} and EDVR~\cite{wang2019edvr} employ the deformable alignment in single- or multi-scale feature levels.
Beyond the existing methods, we propose a new strategy that learns to construct dynamic filters, explicitly incorporating multi-frame information while avoiding the feature alignment and optical-flow computation.

%%%\vspace*{2mm}
\noindent\textbf{Video denoising.} \
Early approaches are mostly patch-based, e.g., V-BM4D~\cite{maggioni2012video} and VNLB~\cite{arias2018video}, which extend from BM3D~\cite{dabov2007image}. 
Later, deep neural networks are explored for the task.
Chen et al.~\cite{chen2016deep} propose the first attempt to video denoising based on RNN.
Vogels et al.~\cite{vogels2018denoising} design a kernel-predicting neural network for denoising Monte-Carlo-rendered sequences. Tassano et al.~\cite{tassano2019dvdnet} propose DVDnet by separating the denoising of a frame into two stages.
More recently, Tassano et al.~\cite{tassano2020fastdvdnet} propose FastDVDnet to eliminate the dependence on motion estimation.
Besides, some recent works focus on blind video denoising, e.g.,~\cite{ehret2019model} and~\cite{claus2019videnn}.

%%%\vspace*{2mm}
\noindent\textbf{Video illumination enhancement.} \
Learning-based low-light image enhancement gains increasing attention recently~\cite{yan2014learning,yan2016automatic,lore2017llnet,cai2018learning,Wang_2019_CVPR,Moran_2020_CVPR,Zero-DCE}. 
Wang et al.~\cite{Wang_2019_CVPR} enhance photos by learning to estimate an illumination map. 
Sean et al.~\cite{Moran_2020_CVPR} learn spatial filters of various types for image enhancement.
Also, unsupervised learning has been explored, e.g., Guo et al.~\cite{Zero-DCE} train a lightweight network to estimate pixel-wise and high-order curves for dynamic range adjustment.
Yet, applying low-light image enhancement methods independently to individual frames will likely cause flickering, thus leading to research on methods for low-light videos, e.g.,~\cite{zhang2016underexposed,lv2018mbllen,jiang2019learning,xue2019video,wang2019enhancing,chen2019seeing}.
Zhang et al.~\cite{zhang2016underexposed} adopt a perception-driven progressive fusion.
Lv et al.~\cite{lv2018mbllen} design a multi-branch network to extract multi-level features for stable enhancement. 
%%Jiang et al.~\cite{jiang2019learning} learn the enhancement mapping for transforming low-light RAW videos to normal-light videos.
%%%Chen et al.~\cite{chen2019seeing} compile a dataset and learned RAW-to-RGB transformation for videos.

\vspace{-0.1in}
\section{Method}

\begin{figure*}[t]
	\begin{center}
		\includegraphics[width=0.8\linewidth]{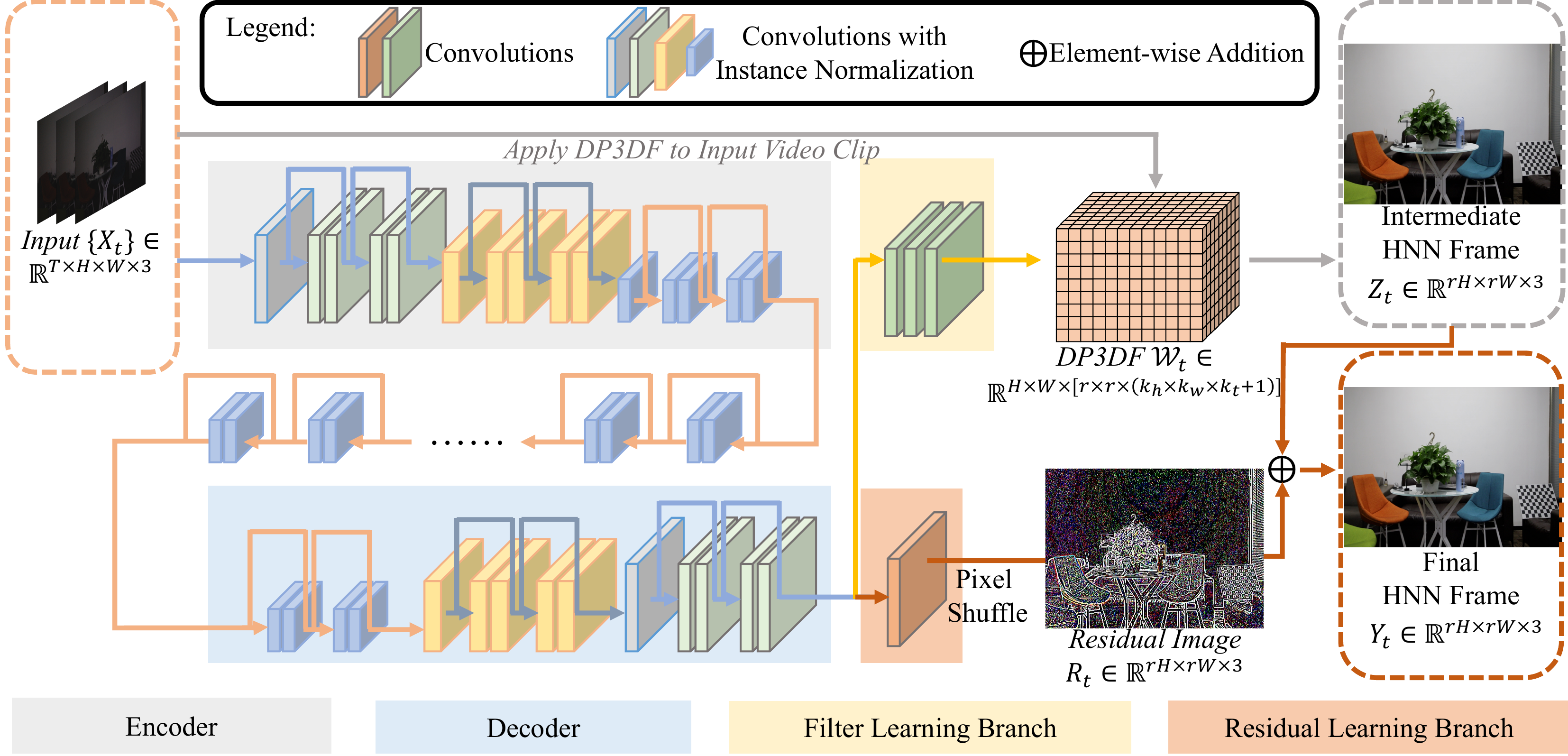}
	\end{center}
	\vspace{-0.2in}
	\caption{
		Overview of our framework.
		The encoder branch (grey area) extracts deep features from network input $\{X_t\}$ and the decoder branch (blue area) produces the output for learning the DP3DF and residual image $R_t$ (the branch in the red area).
		Further, we learn the DF3DF (the branch in the yellow area) for synthesizing the intermediate HNN frame $Z_t$.
		Finally, we refine $Z_t$ using $R_t$ to produce the final output $Y_t$.
		The branch for learning the DP3DF and residual image share the same encoder-and-decoder backbone.
		Also, our DP3DF explicitly exploits adjacent temporal-spatial information around each pixel; see Fig.~\ref{fig:dp3df} for the details of how to apply DP3DF to a video.
	}
	\vspace{-0.2in}
	\label{fig:framework}
\end{figure*}

\subsection{Architecture}
To start, let us denote $\{X_t\}$ as the input LLN frames and $\{Y_t\}$ as the synthesized HNN frames, where $t$ is the time index.
Usually, we train the network with $\{X_t\}$ downsampled from the ground-truth frames $\{\widehat{Y}_t\}$ and we denote $r$ as the downsampling rate.
To obtain realistic and temporally-smooth videos, we consider $N$ frames before and $N$ frames after time $t$ for estimating the target frame $Y_t$:
\begin{equation}
	\small
	Y_t=f(X_{t'}, t'\in [t-N, t+N]).
\end{equation} 
Thus, the shape of the network input is $T \times H \times W \times C$, where $T=2N+1$ and $H$, $W$, $C$ are the height, width, channel size of the input video.
Then, the shape of the output SR frame $Y_t$ shall be $rH \times rW \times C$.
Fig.~\ref{fig:framework} illustrates the network input, synthesized frame, and various components in our framework.
Overall, our framework first synthesizes an intermediate HNN frame $Z_t$, then constructs residual image $R_t$ to refine $Z_t$ to generate the final output $Y_t$.

\begin{figure*}[t]
	\begin{center} 
		\includegraphics[width=0.8\linewidth]{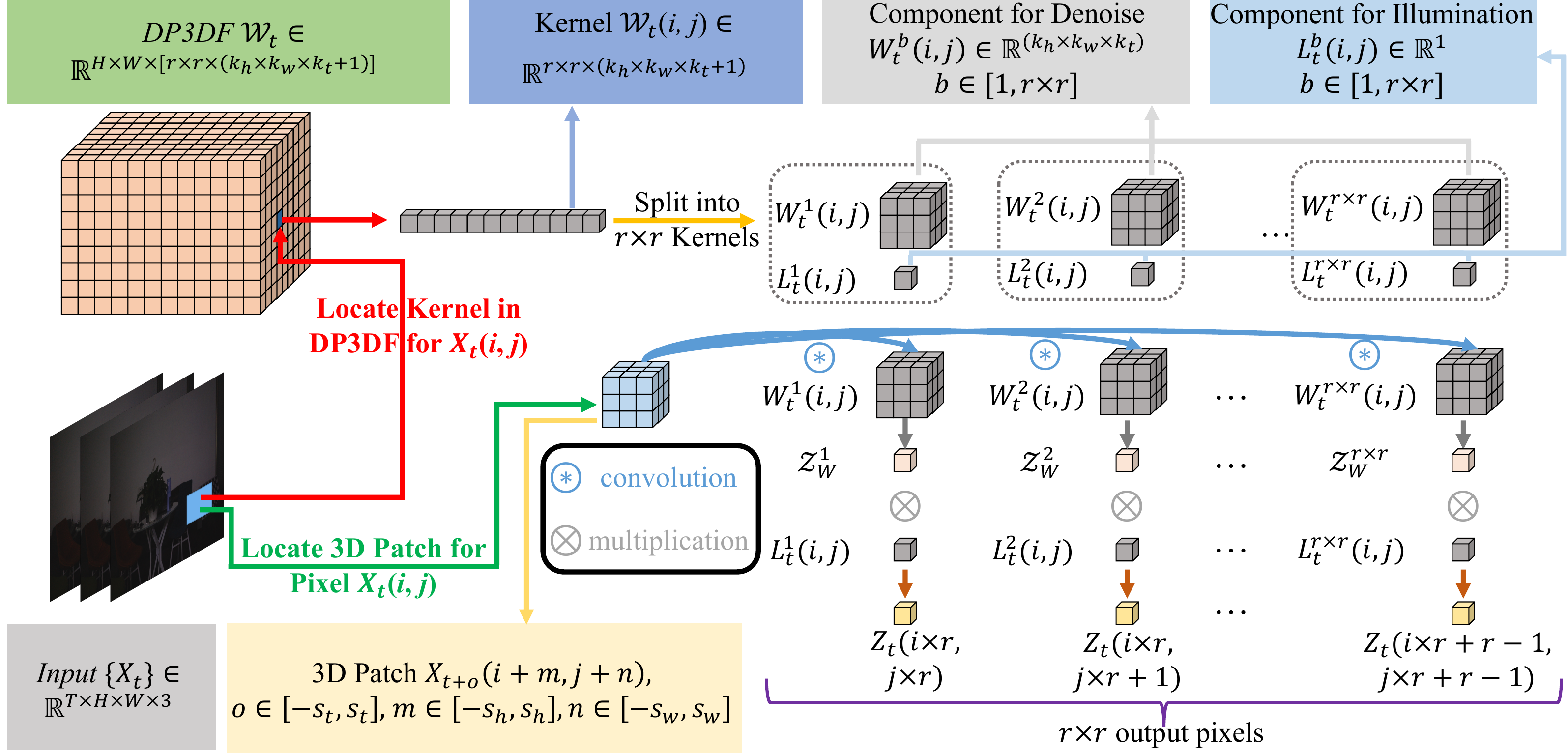}
	\end{center}
	\vspace{-0.2in}
	\caption{Illustrating how we apply the learned DP3DF to process an input video clip.
		For each pixel $X_t(i,j)$, we locate its 3D patch (green arrow) and its associated DP3DF kernel (red arrow), and then make use of the kernel components to process the 3D patch to produce $r \times r$ output pixels (yellow and blue arrows).
	}
	\vspace{-0.15in}
	\label{fig:dp3df}
\end{figure*}

Rather than directly producing the pixels of $Z_t$ (which is HNN), we propose to first learn a new parametric representation called DP3DF.
To complete the mapping from $\{X_t\}$ to $Z_t$, DP3DF has the shape of $H \times W \times [r\times r \times (k_h \times k_w \times k_t+1)]$, where $k_h$, $k_w$, and $k_t$ are the dimensions (height, width, and time, respectively) of a 3D volume covered by DP3DF at each pixel in the network input and the ``+1'' is an additional component for illumination enhancement.
Each pixel has $r\times r$ DP3DF kernels, each of size $k_h \times k_w \times k_t+1$; see Fig.~\ref{fig:dp3df}.
For the enhancement of $X_t(p)$, where $p$ denotes a pixel location, we sample a volume of $k_h \times k_w \times k_t$ pixels around $p$ in $X_t$ and then use the learned $r\times r$ kernels to produce $r \times r$ pixels for the original pixel at $p$.
Besides, we normalize the $k_h \times k_w \times k_t$ elements in each kernel to be a sum of one for promoting smoothness in the results and suppressing the noise.
Further, for the kernel prediction at each pixel, the additional ``one'' dimension is predicted for illumination adjustment in dark areas.
The details of the kernels will be discussed in the next subsections.

\vspace{-0.1in}
\subsection{DP3DF}
\noindent\textbf{Formulation.} \
Our network predicts DP3DF $\mathcal{W}_t$ from $\{X_t\}$ and the filter learning branch output; see Fig.~\ref{fig:framework}.
Specifically, the DP3DF kernel $\mathcal{W}_t(p)$ associates with pixel $X_t(p)$ in $X_t$.
Each $\mathcal{W}_t(p)$ can be decomposed into $r\times r$ kernels.
Each kernel has shape $k_h \times k_w \times k_t+1$ and can be decomposed into two parts:
$W^b_t(p) \in \mathbb{R}^{k_h \times k_w \times k_t}$ (weights for SR and denoising) and
$L^b_t(p) \in \mathbb{R}^{1}$ (weight for luminance adjustment), where $b\in[1, r\times r]$.
Suppose $s_h=\frac{k_h-1}{2}$, $s_w=\frac{k_w-1}{2}$, $s_t=\frac{k_t-1}{2}$, $p=(i,j)$, the upsampled $r\times r$ pixels in $Z_t$ can be predicted as
\begin{equation}
	%\large
	\begin{split}
		&\resizebox{0.8\linewidth}{!}{$\mathcal{Z}^{r_1\times r + r_2}_{W}=\sum \limits_{m=-s_h}^{s_h} \sum \limits_{n=-s_w}^{s_w} \sum \limits_{o=-s_t}^{s_t} W(m,n,o) \times X_{t+o}(i+m, j+n),$} 
		\\
		&\resizebox{0.7\linewidth}{!}{$Z_t(i\times r+r_1, j\times r+r_2) = 	\mathcal{Z}^{r_1\times r + r_2}_{W} \times L^{r_1\times r +r_2}_{t}(i, j),$}
	\end{split}
\end{equation}
where $r_1 \in \{0,1,...,r-1\}$ and $r_2 \in \{1,2,...,r\}$, which together iterate over the $r \times r$ kernels in $W_t(p)$, and
$W(m,n,o)$ denotes $W^{r_1\times r +r_2}_{t}(i, j)[m+s_h, n+s_w, o+s_t]$.
Especially, the elements in $W^b_t(p)$ are normalized through Softmax, summing to one, whereas the elements in $L^b_t(p)$ are processed with the activation function of Sigmoid and we take reciprocals.
The convolution with $W^b_t(p)$ gives an effect of spatial-temporal smoothing and helps achieve denoising.
On the other hand, the multiplication with $L^b_t(p)$ adjusts the illumination and enhances the dark areas in the input frame.
Also, the resulting $r\times r$ pixels produce a high-resolution frame from the low-resolution one.
Thus, we generate the DP3DF locally and dynamically, considering the spatiotemporal neighborhood of each pixel. 

Our model has several significant advantages.
First, different operations can be formulated within a single coherent representation and can be learned jointly to enhance the performance.
Second, such a representation allows us to propagate information, both spatially and temporally.

%%\vspace*{2mm}
\noindent\textbf{Implementation.} \
To learn the DP3DF, we adopt a network of an encoder-and-decoder structure.
As shown in Fig.~\ref{fig:framework}, the encoder has two downsampling layers, each with several residual blocks~\cite{he2016deep}. These residual blocks can extract relevant features in each layer and use an instance normalization to reduce the gap between different types of videos.
Then, we pass the features from the encoder through several residual blocks to produce the input feature of the decoder.
Subsequently, the decoder adopts a pixel shuffle~\cite{shi2016real} for upsampling.

\begin{figure*}[t]
	\begin{center} 
		\includegraphics[width=0.8\linewidth]{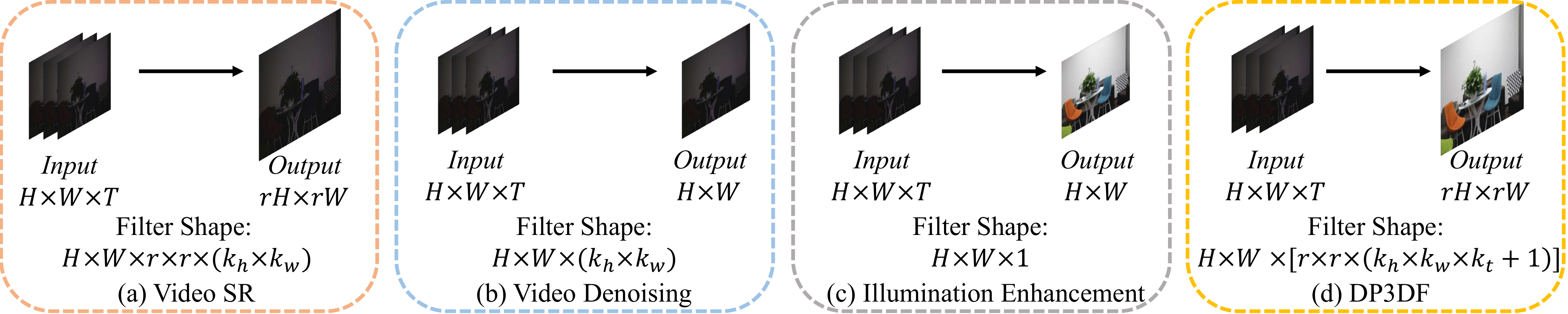}
	\end{center}
	\vspace{-0.2in}
	\caption{Illustrating dynamic filters implemented for different tasks. Our DP3DF (d) is a generalized filter that combines the properties of the dynamic filters for various tasks (a)-(c). Our DP3DF simultaneously achieves SR, denoising, and illumination enhancement, and is able to explicitly incorporate adjacent frame information, unlike the existing dynamic filter methods.}
	\vspace{-0.25in}
	\label{fig:visual}
\end{figure*}

%%\vspace*{2mm}
\noindent\textbf{Connection with existing dynamic filters.} \
\label{connection}
As illustrated in Fig.~\ref{fig:visual}, we design our DP3DF to be a generalized filter that combines the properties of dynamic filters for different tasks.  
The dynamic filters in video SR, denoising, and illumination enhancement are special cases of DP3DF. Especially, the dynamic filters~\cite{jo2018deep} in SR predict $r\times r$ kernels per pixel with a kernel size of $k_h\times k_w$; existing denoising methods~\cite{mildenhall2018burst,xia2020basis,xia2021deep} predict one kernel of size $k_h\times k_w$ per pixel to smooth the neighborhood area, whereas existing illumination enhancement methods, e.g.,~\cite{Wang_2019_CVPR}, adopt the Retinex theory to predict one value per pixel.
If we reduce the 3D kernel shape ($k_h \times k_w \times k_t$) to 2D ($k_h\times k_w$) and remove the dimension for illumination enhancement, we trim down the DP3DF into a dynamic filter for regular SR tasks.
If we reduce the number of predicted kernels from $r\times r$ to one and remove the dimension for illumination enhancement, we turn the DP3DF into a filter for denoising.
Lastly, if we reduce the number of predicted kernels from $r\times r$ to one and remove the 3D kernel, we trim down the DP3DF into a filter for regular illumination enhancement.

\vspace{-0.1in}
\subsection{Residual Learning}
To further enhance the performance, we adopt a residual learning branch (see the red area in Fig.~\ref{fig:framework}) to learn a residual image $R_t$ for enriching the final output with high-frequency details.
Importantly, the residual image $R_t$ is produced from multiple input frames $\{X_t\}$ rather than a single input frame, so sharing the same encoder-decoder structure with the main branch for predicting the DP3DF allows us to reduce the computational overhead.
Finally, we combine the intermediate HNN frame $Z_t$ with the learned residual image $R_t$ to produce final output frame $Y_t$.

\vspace{-0.1in}
\subsection{Loss Function}
%%The overall loss for training our framework has the following three parts.
The overall loss has the following three parts.

%%\vspace*{2mm}
\noindent\textbf{(i) Reconstructing $Z_t$.} \
First, we define an $L_2$ loss term for obtaining an accurate prediction of $Z_t$ with the DP3DF:
\begin{equation}
	\small
	\mathcal{L}_r=\Vert Z_t-\widehat{Y}_t\Vert,
\end{equation}
where $\Vert \Vert$ is the $L_2$ norm, and all pixel channels in ground truth $\widehat{Y}_t$ and $Z_t$ are normalized to [0, 1]. Such clip operation is effective for the training of illumination enhancement, eliminating invalid colors that are beyond the gamut and avoiding mistakenly darkening the underexposed regions.

%%\vspace*{2mm}
\noindent\textbf{(ii) Residual learning branch.} \
Like $Z_t$, we define another reconstruction loss for the residual learning branch to generate the final output $Y_t$ from $Z_t$ and $R_t$:
\begin{equation}
	\small
	\mathcal{L}_{e}=\Vert Y_t-\widehat{Y}_t\Vert.
\end{equation}

%%\vspace*{2mm}
\noindent\textbf{(iii) Smoothness loss.} \
Many works employ the smoothness prior for illumination enhancement,~e.g.,~\cite{li2014single,Wang_2019_CVPR}, by assuming the illumination is locally smooth.
Harnessing this prior in our framework has two advantages.
It helps to not only reduce overfitting and improve the network’s generalizability but also enhance the image contrast.
For adjacent pixels, say $p$ and $q$, with similar illumination values in a video frame, their contrast in the enhanced frame should be small; and vice versa.
So, we define the smoothness loss on the predicted $L_t^m$ as
\begin{equation}
	\resizebox{0.7\linewidth}{!}{$\mathcal{L}_s=\sum_t \sum_m \sum_p [v_t^p \times [\partial_x L_t^m(p)]^2 + u_t^p \times [\partial_y L_t^m(p)]^2]$},
\end{equation}
where $\partial_x$ and $\partial_y$ are partial derivatives in horizontal and vertical directions, respectively, for the predicted $L_t^m$; $v_t^p$ and $u_t^q$ are spatially-varying
(per-channel) smoothness weights expressed as
\begin{equation}
	%\begin{split}
	\small
	v_t^p=(\Vert \partial_x \mathcal{X}_t(p) \Vert ^{1.2} + \epsilon )^{-1} \ \text{and} \ u_t^p=(\Vert \partial_y \mathcal{X}_t(p) \Vert ^{1.2}+ \epsilon)^{-1},
	%\end{split}
\end{equation}
where $\mathcal{X}_t$ is the logarithmic image of $X_t$; 
and $\epsilon$ is a small constant (set to 0.0001) to prevent division by zero.

%%\vspace*{2mm}
\noindent\textbf{Overall loss.} \
%%We combine the three losses into $\mathcal{L}$ to train the network:
The overall loss $\mathcal{L}$  is
\begin{equation}
	\small
	\mathcal{L}=\lambda_1 \mathcal{L}_r + \lambda_2 \mathcal{L}_s + \lambda_3 \mathcal{L}_e,
\end{equation}
where $\lambda_1$, $\lambda_2$ and $\lambda_3$ are the loss weights.
%\textbf{\em We will release code and trained models upon the publication of this work.}

%\vspace{-0.1in}
\section{Experiments}

%\vspace{-0.1in}
\subsection{Datasets}
We perform our evaluation on two public datasets with indoor and outdoor real-world videos: SMID~\cite{chen2019seeing} and SDSD~\cite{sdsd}.
The videos in SMID are captured as static videos, in which the ground truths are obtained with a long exposure and the signal-to-noise ratio of the videos under the dark environment is extremely low.
In this work, we explore the mapping from LLN to HNN frames in the sRGB domain. Thus, we follow the script provided by SMID~\cite{chen2019seeing} to convert the low-light videos from the RAW domain to the sRGB domain using rawpy's default ISP.
On the other hand, SDSD is a dynamic video dataset collected through an electromechanical equipment, containing indoor and outdoor subsets.
%It has two subsets that contain indoor and outdoor videos.
Also, we follow the official train-test split of SMID and SDSD.

\vspace{-0.1in}
\subsection{Implementation}
We empirically set $k_h$$=$$k_w$$=$$k_t$$=$$3$ and number of frames $T$$=$$3$.
Experiments on all datasets were conducted on the same network structure, whose backbone is an encoder-and-decoder structure; see Fig.~\ref{fig:framework}.
The encoder has three down-sampling layers with 64, 128, 256 channels, while the decoder has three up-sampling layers with 256, 128, 64 channels.
The branches for learning the DP3DF and residual have three and one convolution layers, respectively.
The shape of the DP3DF kernel is $r$$\times$$r$$\times$$(k_h$$\times$$k_w$$\times$$k_t$$+$$1)$, in which the first $r$$\times$$r$$\times$$(k_h$$\times$$k_w$$\times$$k_t)$ dimensions contain $r$$\times$$r$ parts for denoising and the remaining dimensions contain $r$$\times$$r$ parts for illumination enhancement.

We train all modules end-to-end with the learning rate initialized as 4e-4 for all layers (adapted by the cosine learning scheduler); scale factor $r=4$; batch size = 16; and patch size = $64\times64$. The patches are cropped randomly from the 
down-sampled low-resolution frame.
We use Kaiming Initialization~\cite{he2015delving} to initialize the weights and Adam~\cite{kingma2014adam} for training with momentum set to 0.9.
Besides, we perform data augmentation by random rotation of $90^{\circ}$/$180^{\circ}$/$270^{\circ}$ and horizontal flip.
We implement our method using Python 3.7.7 and PyTorch 1.2.0~\cite{paszke2019pytorch}, and ran all experiments on one NVidia TITAN XP GPU.
PSNR and SSIM~\cite{wang2004image} are adopted for quantitative evaluation.

\begin{table}[!t]
	\centering
	%\large
	\resizebox{1.0\linewidth}{!}{
		\begin{tabular}{c|p{1.5cm}<{\centering}p{1.5cm}<{\centering}|p{1.5cm}<{\centering}p{1.5cm}<{\centering}|p{1.5cm}<{\centering}p{1.5cm}<{\centering}}
			\toprule[1pt]
			& \multicolumn{2}{c|}{SMID} &\multicolumn{2}{c|}{SDSD Indoor}&\multicolumn{2}{c}{SDSD Outdoor}\\
			\hline
			Methods & PSNR & SSIM& PSNR & SSIM& PSNR & SSIM \\
			\hline
			Ours w/o Temporal &23.67 &0.69 &25.49&0.83&24.98&0.75  \\
			Ours w/o Spatial&22.84 &0.63&24.87&0.78&24.02&0.71  \\
			Ours w/o Residual &25.44 &0.71 &27.01&0.83&25.69&0.76  \\
			\hline 
			Ours &\textbf{25.73} &\textbf{0.73} &\textbf{27.11}&\textbf{0.85}&\textbf{25.80}&\textbf{0.77}  \\
			\bottomrule[1pt]
	\end{tabular}}
	\vspace{-0.15in}
	\caption{The quantitative evaluation in the ablation study.}
	\label{tab:abla}
	%\vspace{-0.15in}
\end{table}

\begin{figure}[t]
	\centering
	\captionsetup[subfigure]{labelformat=empty}
	\begin{subfigure}[c]{0.23\textwidth}
		\centering
		\includegraphics[width=1.6in]{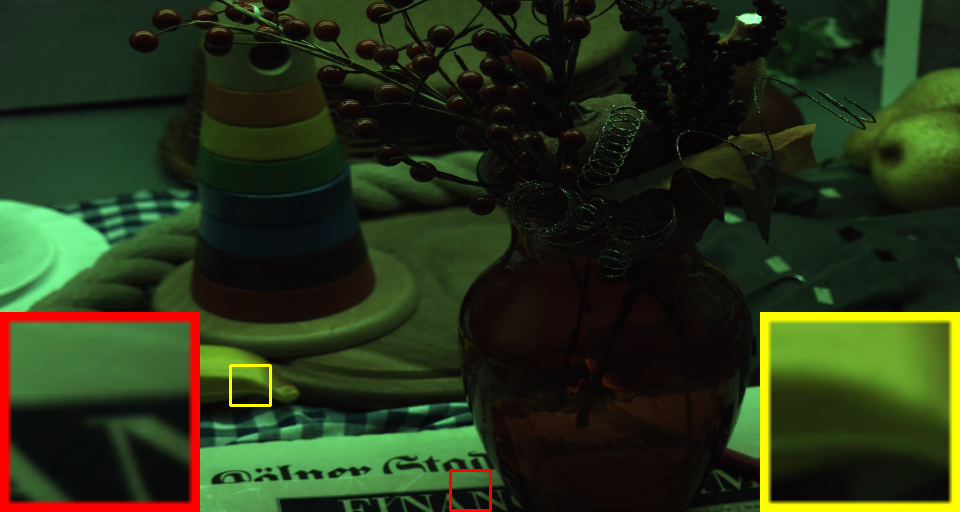}
		%\vspace{-0.5em}
		\caption{(a) Input}
		%	{\scriptsize(PSNR: 11.06, SSIM: 0.48)}
	\end{subfigure}
	\begin{subfigure}[c]{0.23\textwidth}
		\centering
		\includegraphics[width=1.6in]{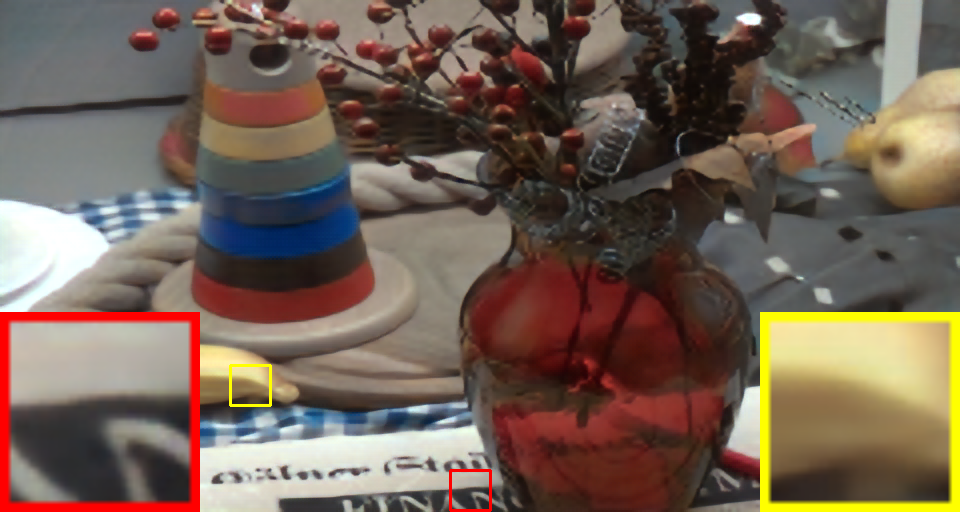}
		%\vspace{-0.5em}
		\caption{(b) Ours w/o Temporal}
		%	{\scriptsize(PSNR: 23.91, SSIM: 0.80)}
	\end{subfigure}
	\vspace{0.1em}\\
	\begin{subfigure}[c]{0.23\textwidth}
		\centering
		\includegraphics[width=1.6in]{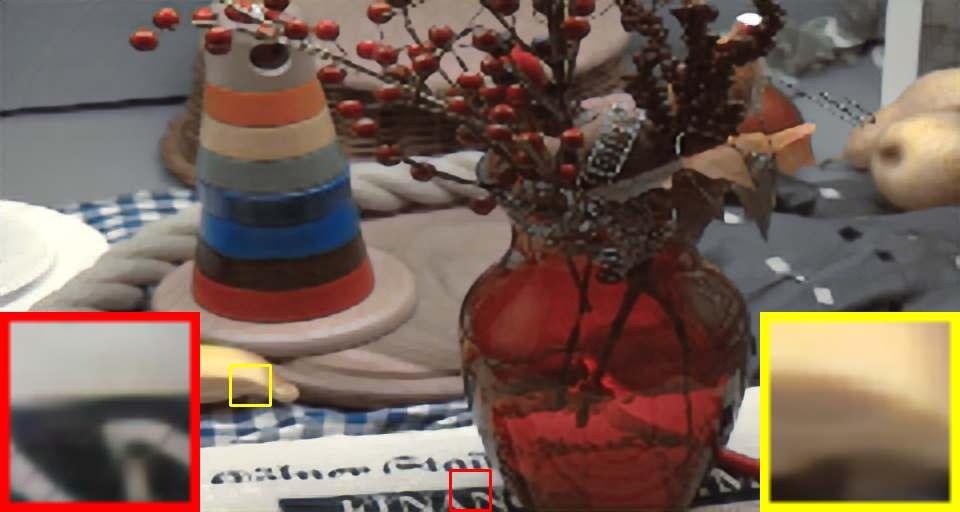}
		%\vspace{-0.5em}
		\caption{(c) Ours w/o Spatial}
		%	{\scriptsize(PSNR: 23.57, SSIM: 0.79)}
	\end{subfigure}
	\begin{subfigure}[c]{0.23\textwidth}
		\centering
		\includegraphics[width=1.6in]{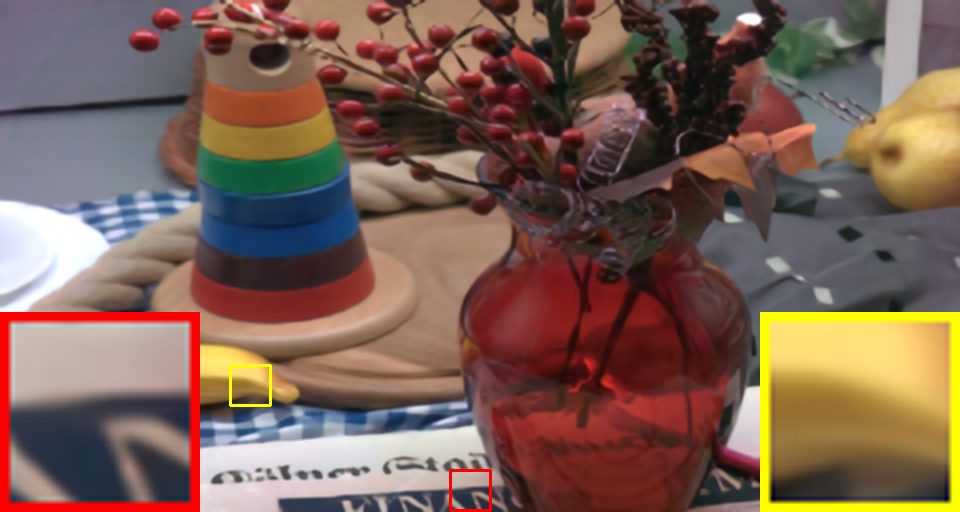}
		%\vspace{-0.5em}
		\caption{(d) Ours}
		%	{\scriptsize(PSNR: \textbf{26.17}, SSIM: \textbf{0.83})}
	\end{subfigure}
	\vspace{-0.1in}
	\caption{Example visual samples in the ablation study.}
	\label{fig:abla}
	\vspace{-0.25in}
\end{figure}

\vspace{-0.1in}
\subsection{Ablation Study}
We evaluate the major components in DP3DF on three ablated cases:
(i) ``w/o Temporal'' removes the property of the 3D filters by ignoring the temporal dimension and filtering only in the spatial dimensions;
(ii) ``w/o Spatial'' removes the spatial dimensions in DP3DF and applies filters only in the temporal dimension; and
(iii) ``w/o Residual'' removes the branch of residual learning.

Table~\ref{tab:abla} summarizes the results, showing that all ablated cases are weaker than our full method.
Especially, ``w/o Temporal'' does not have the ability to incorporate information from the adjacent time frames and ``w/o Spatial'' cannot obtain information from the adjacent pixels, thereby both having weaker performance.
These two cases show the necessity of considering both the temporal and spatial dimensions in our 3D filter.
Though ``w/o residual'' leverages multiple frames as DP3DF, our full model still consistently achieves better results.
Further, Fig.~\ref{fig:abla} shows some visual samples, revealing the apparent degradation caused by removing different components in our 3D filter. Fine details and textures are better reconstructed using our full model.

\begin{figure*}[t]
	\centering
	\begin{subfigure}[c]{0.24\textwidth}
		\centering
		\includegraphics[width=1.6in]{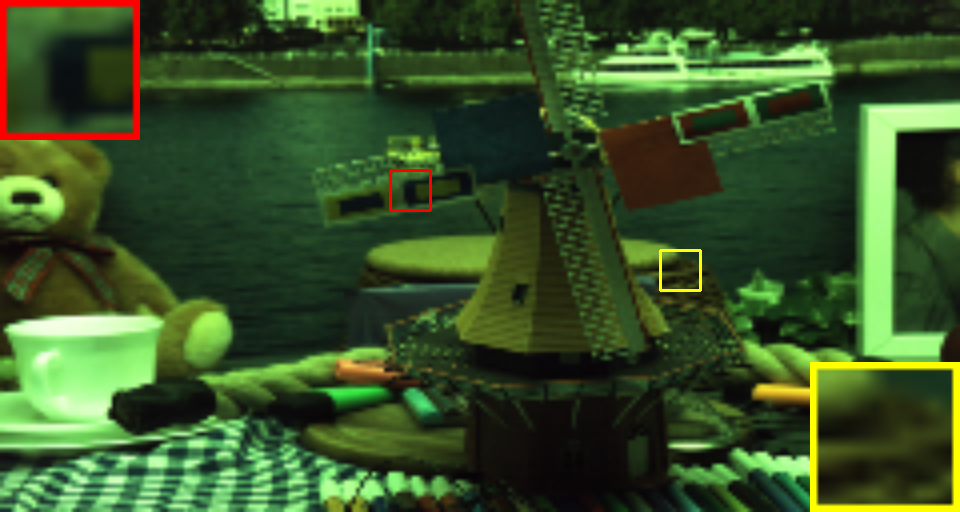}
		%\vspace{-1.5em}
		\caption{Input {\scriptsize PSNR: 9.58, SSIM: 0.50}}
	\end{subfigure}
	\begin{subfigure}[c]{0.24\textwidth}
		\centering
		\includegraphics[width=1.6in]{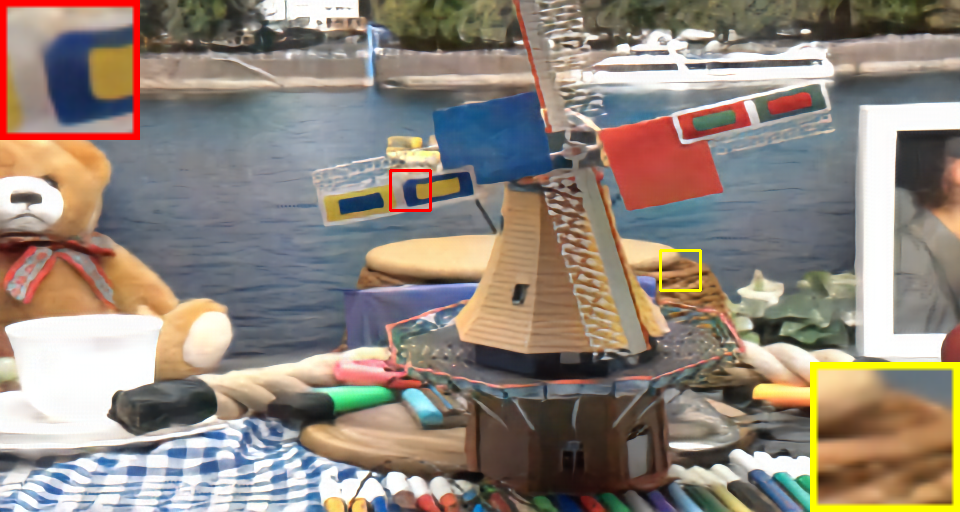}
		%\vspace{-1.5em}
		\caption{RBPN {\scriptsize PSNR: 23.29, SSIM: 0.78}}
	\end{subfigure}
	\begin{subfigure}[c]{0.24\textwidth}
		\centering
		\includegraphics[width=1.6in]{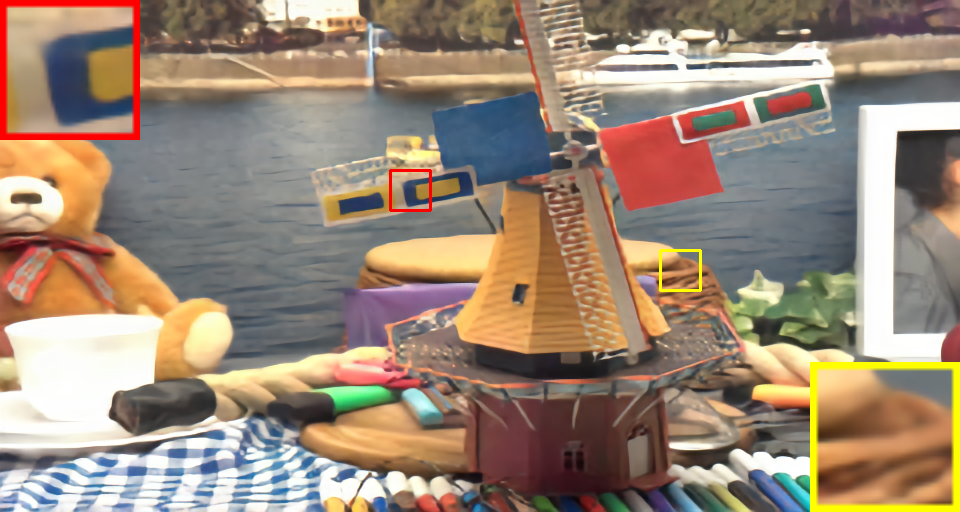}
		%\vspace{-1.5em}
		\caption{Zooming {\scriptsize PSNR: 23.52, SSIM: 0.78}}
	\end{subfigure}
	\begin{subfigure}[c]{0.24\textwidth}
		\centering
		\includegraphics[width=1.6in]{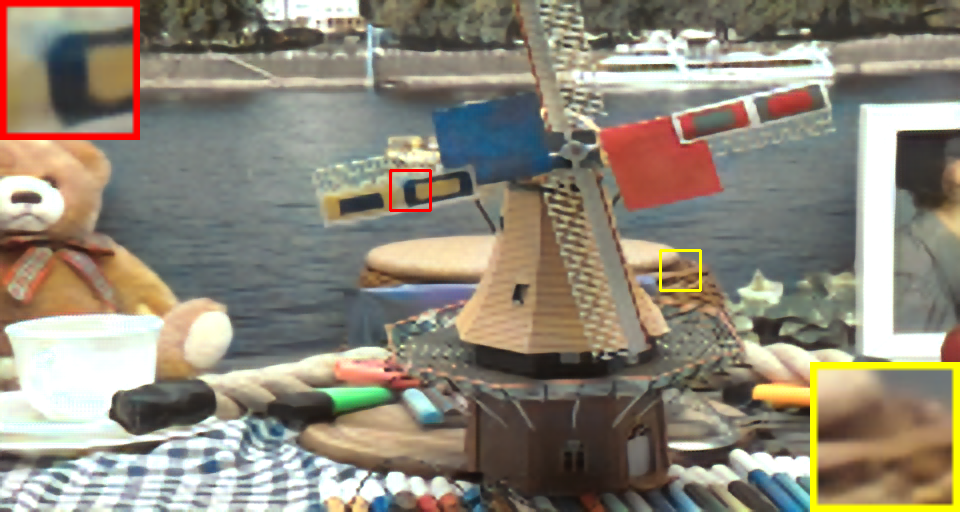}
		%\vspace{-1.5em}
		\caption{TGA {\scriptsize PSNR: 22.36, SSIM: 0.75}}
	\end{subfigure} 
	\\
	\begin{subfigure}[c]{0.24\textwidth}
		\centering
		\includegraphics[width=1.6in]{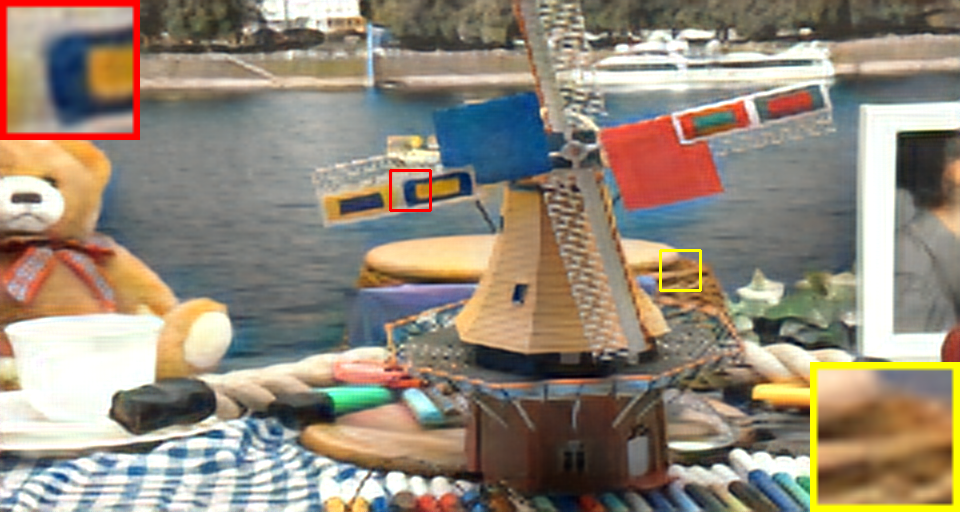}
		%\vspace{-1.5em}
		\caption{TDAN {\scriptsize PSNR: 22.92, SSIM: 0.75}}
	\end{subfigure} 
	\begin{subfigure}[c]{0.24\textwidth}
		\centering
		\includegraphics[width=1.6in]{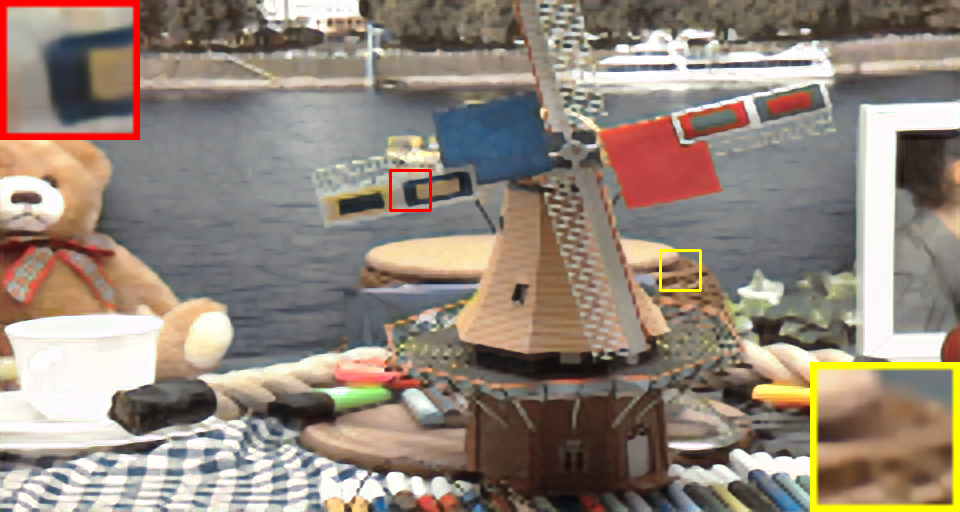}
		%\vspace{-1.5em}
		\caption{ToFlow, {\scriptsize PSNR: 22.27, SSIM: 0.75}}
	\end{subfigure}
	\begin{subfigure}[c]{0.24\textwidth}
		\centering
		\includegraphics[width=1.6in]{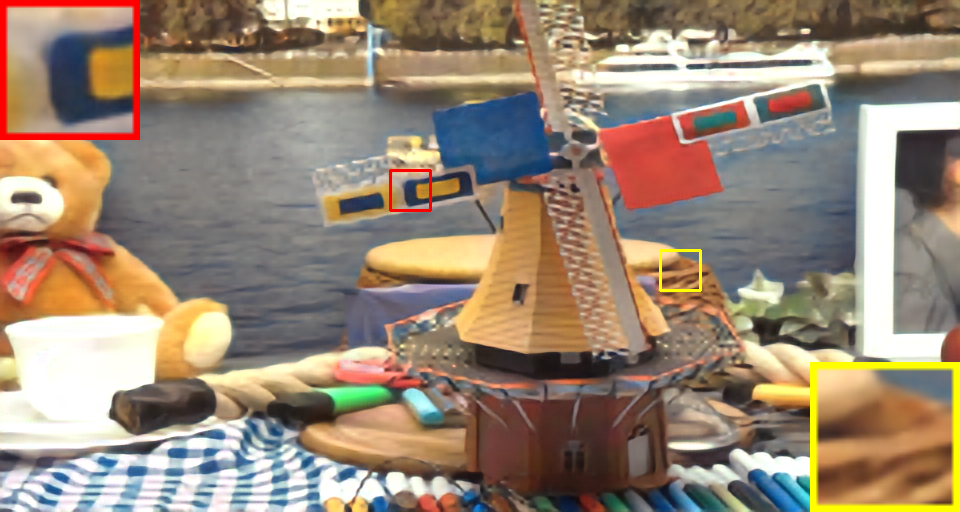}
		%\vspace{-1.5em}
		\caption{EDVR {\scriptsize PSNR: 23.44, SSIM: 0.77}}
	\end{subfigure}
	\begin{subfigure}[c]{0.24\textwidth}
		\centering
		\includegraphics[width=1.6in]{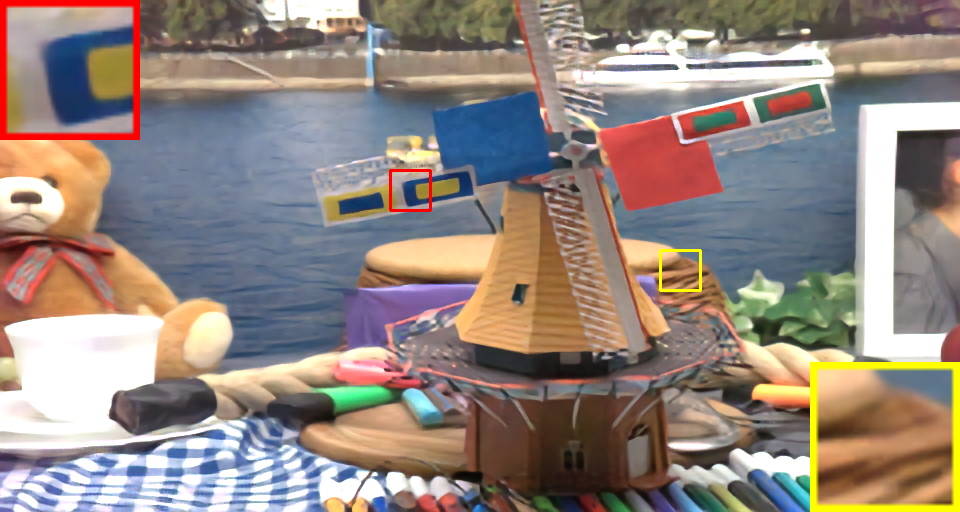}
		%\vspace{-1.5em}
		\caption{Ours {\scriptsize PSNR: \textbf{24.54}, SSIM: \textbf{0.79}}}
	\end{subfigure}
	\vspace{-0.15in}
	\caption{Qualitative comparison on SMID. 
		Our result contains sharper details and more vivid colors. 
		Please zoom to view.}
	\vspace{-0.2in}
	\label{fig:cmp_smid}
\end{figure*}

\begin{table}[t]
	\centering
	\resizebox{1.0\linewidth}{!}{
		\begin{tabular}{l|p{1.5cm}<{\centering}p{1.5cm}<{\centering}|p{1.5cm}<{\centering}p{1.5cm}<{\centering}|p{1.5cm}<{\centering}p{1.5cm}<{\centering}}
			\toprule[1pt]
			& \multicolumn{2}{c|}{SMID} &\multicolumn{2}{c|}{SDSD Indoor}&\multicolumn{2}{c}{SDSD Outdoor}\\
			\hline
			Methods & PSNR & SSIM& PSNR & SSIM& PSNR & SSIM \\
			\hline
			BasicVSR &21.78&0.62&20.72&0.71&20.91&0.70\\
			BasicVSR++&22.48&0.65&21.02&0.75&21.31&0.72\\
			IconVSR &21.99&0.63&20.94&0.73&20.89&0.71\\
			RBPN &24.87 &0.72 &23.47&0.80&22.46&0.74  \\
			Zooming&24.89 &0.71&26.32&0.84&22.05&0.72 \\
			TGA &23.40 &0.67&23.92&0.76&23.83&0.74 \\
			TDAN&24.65 &0.70&24.00&0.80&22.57&0.74 \\
			PFNL&20.85 &0.60&23.19&0.82&23.31&0.72 \\
			ToFlow &23.08 &0.66&21.82&0.76&22.07&0.71 \\
			EDVR & 24.50&0.70&25.00&0.83&23.37&0.75 \\
			\hline 
			Ours &\textbf{25.73} &\textbf{0.73} &\textbf{27.11}&\textbf{0.85}&\textbf{25.80}&\textbf{0.77}  \\
			\bottomrule[1pt]
	\end{tabular}}
	\vspace{-0.15in}
	\caption{Quantitative comparison with various SOTA SR methods on the SMID and SDSD datasets.}
	%\huge
	\label{comparison1}
	\vspace{-0.15in}
\end{table}

\begin{table}[!t]
	\centering
	%\huge	
	\large
	\resizebox{1.0\linewidth}{!}{
		\begin{tabular}{l|p{1.35cm}<{\centering}p{1.35cm}<{\centering}|p{1.35cm}<{\centering}p{1.35cm}<{\centering}|p{1.35cm}<{\centering}p{1.35cm}<{\centering}}
			\toprule[1pt]
			& \multicolumn{2}{c|}{SMID} &\multicolumn{2}{c|}{SDSD Indoor}&\multicolumn{2}{c}{SDSD Outdoor}\\
			\hline
			Methods & PSNR & SSIM & PSNR & SSIM & PSNR & SSIM \\
			\hline
			FastDVDnet+Zooming &25.22&0.71&26.82&0.80&22.93&0.72 \\
			FastDVDnet+TGA &23.97&0.70&24.15&0.78&24.11&0.74 \\
			FastDVDnet+TDAN &24.95&0.71&24.30&0.76&23.55&0.70 \\
			TCE+Zooming &24.34&0.67&25.74&0.77&22.15&0.69 \\
			TCE+TGA &23.07&0.66&23.65&0.72&23.48&0.70 \\
			TCE+TDAN &24.12&0.68&23.69&0.71&23.01&0.67 \\
			
			FastDVDnet+TCE+Zooming&23.97&0.74&26.54&0.81&24.41&0.73 \\
			FastDVDnet+TCE+TGA &24.57 &\textbf{0.76}&25.31&0.78&25.01&0.75 \\
			FastDVDnet+TCE+TDAN &24.00 &0.70&25.89&\textbf{0.87}&23.81&0.71 \\
			TCE+FastDVDnet+Zooming&23.73 &0.70&26.01&0.79&23.69&0.74 \\
			TCE+FastDVDnet+TGA &24.66 &0.68&24.70&0.77&24.88&0.72 \\
			TCE+FastDVDnet+TDAN &24.21 &0.71&24.88&0.81&23.35&0.70 \\
			\hline 
			Ours &\textbf{25.73} &0.73 &\textbf{27.11}&0.85&\textbf{25.80}&\textbf{0.77}  \\
			\bottomrule[1pt]
	\end{tabular}}
	\vspace{-0.15in}
	\caption{Comparison with baselines that combine SOTA video SR, denoise, and illumination enhancement networks.}
	\label{comparison2}
	\vspace{-0.15in}
\end{table}

\begin{figure}[t]
	\begin{center} 
		\includegraphics[width=1.0\linewidth]{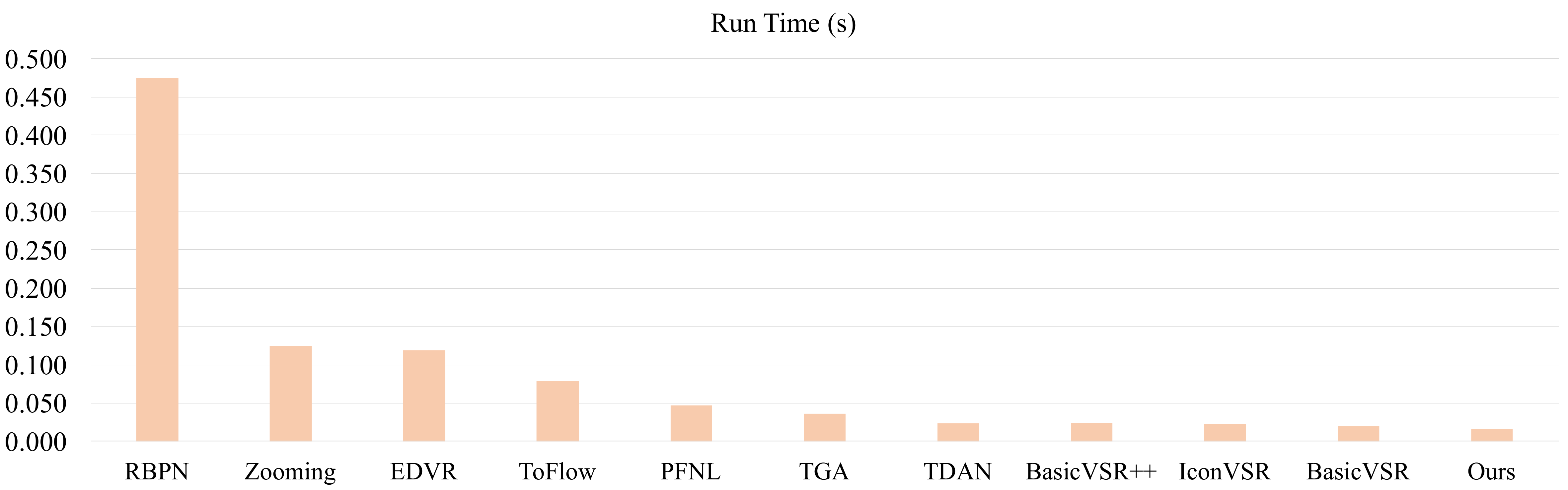}
	\end{center}
	\vspace{-0.2in}
	\caption{Quantitative comparisons between our framework and existing SOTA video SR methods in terms of method run time on input images of 960$\times$512}
	\vspace{-0.25in}
	\label{comparison-speed}
\end{figure}

\vspace{-0.1in}
\subsection{Comparison}

\begin{figure*}[t]
	\centering
	\begin{subfigure}[c]{0.24\textwidth}
		\centering
		\includegraphics[width=1.6in]{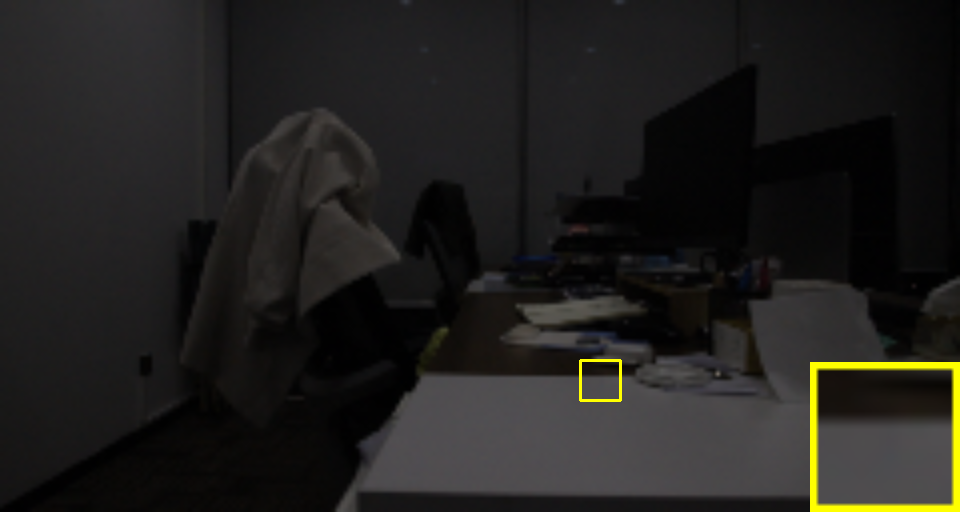}
		%\vspace{-1.5em}
		\caption{Input {\scriptsize PSNR: 7.58, SSIM: 0.35}}
	\end{subfigure}
	\begin{subfigure}[c]{0.24\textwidth}
		\centering
		\includegraphics[width=1.6in]{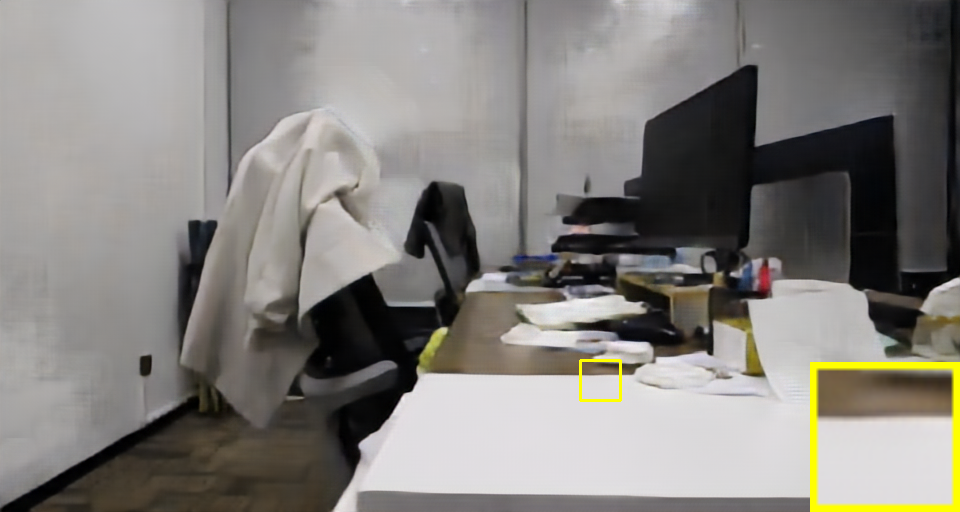}
		%\vspace{-1.5em}
		\caption{RBPN {\scriptsize PSNR: 25.41, SSIM: 0.88}}
	\end{subfigure}
	\begin{subfigure}[c]{0.24\textwidth}
		\centering
		\includegraphics[width=1.6in]{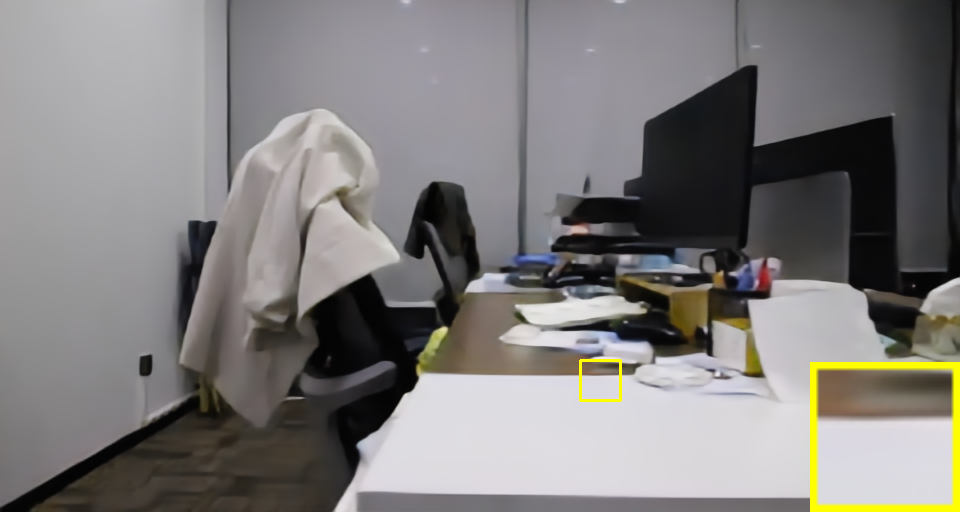}
		%\vspace{-1.5em}
		\caption{Zooming {\scriptsize PSNR: 26.40, SSIM: 0.88}}
	\end{subfigure}
	\begin{subfigure}[c]{0.24\textwidth}
		\centering
		\includegraphics[width=1.6in]{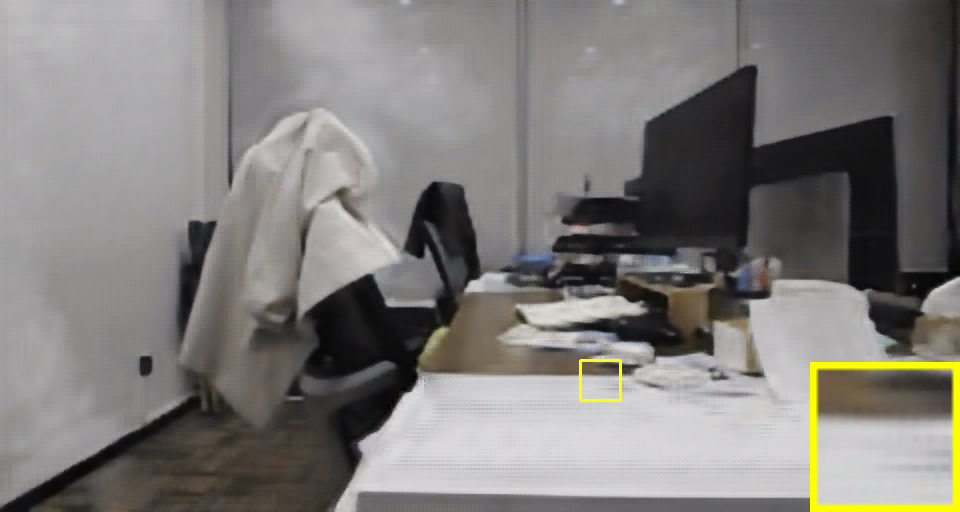}
		%\vspace{-1.5em}
		\caption{TGA {\scriptsize PSNR: 24.98, SSIM: 0.84}}
	\end{subfigure} \\
	\begin{subfigure}[c]{0.24\textwidth}
		\centering
		\includegraphics[width=1.6in]{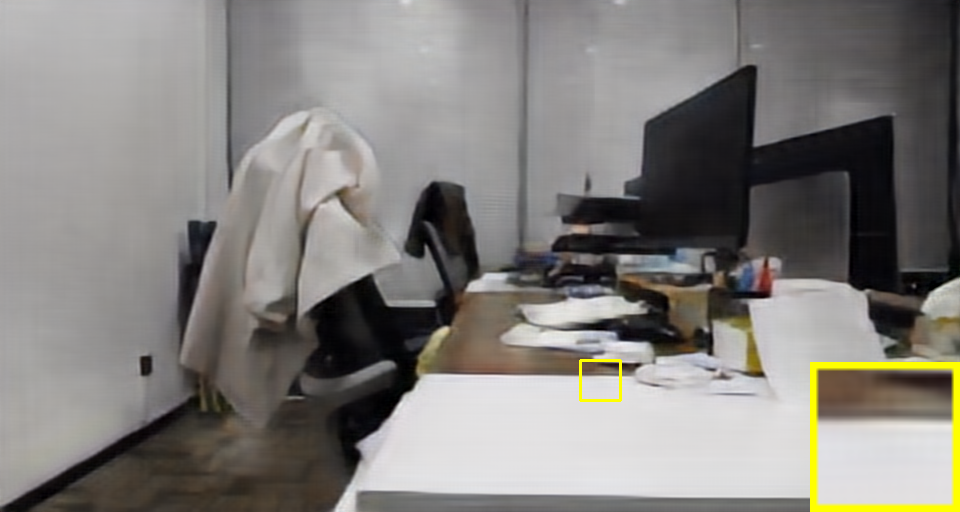}
		%\vspace{-1.5em}
		\caption{TDAN {\scriptsize PSNR: 25.02, SSIM: 0.87}}
	\end{subfigure} 
	\begin{subfigure}[c]{0.24\textwidth}
		\centering
		\includegraphics[width=1.6in]{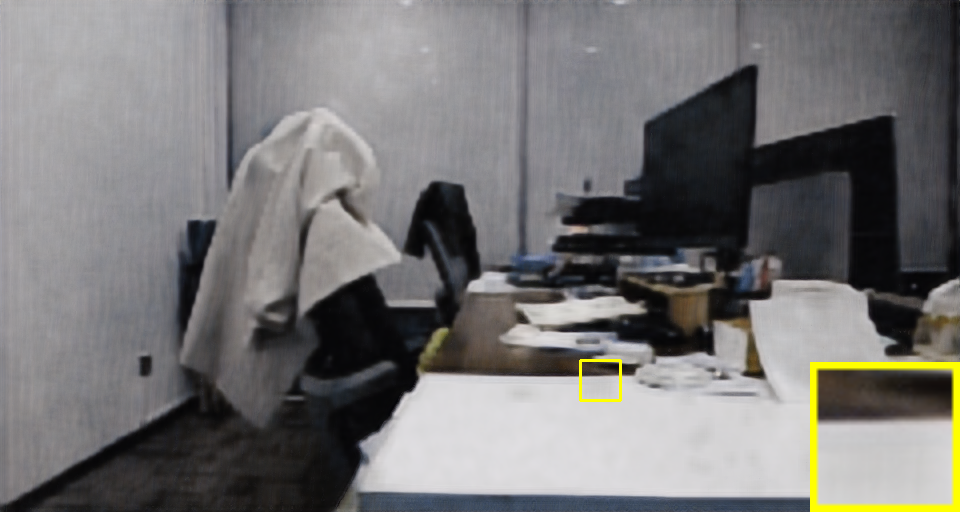}
		%\vspace{-1.5em}
		\caption{ToFlow {\scriptsize PSNR: 22.69, SSIM: 0.83}}
	\end{subfigure}
	\begin{subfigure}[c]{0.24\textwidth}
		\centering
		\includegraphics[width=1.6in]{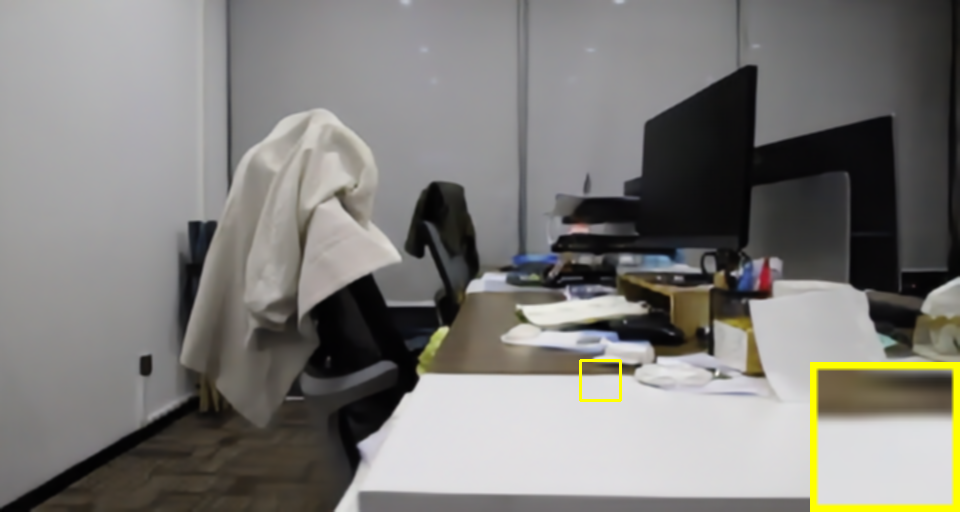}
		%\vspace{-1.5em}
		\caption{EDVR {\scriptsize PSNR: 26.27, SSIM: 0.88}}
	\end{subfigure}
	\begin{subfigure}[c]{0.24\textwidth}
		\centering
		\includegraphics[width=1.6in]{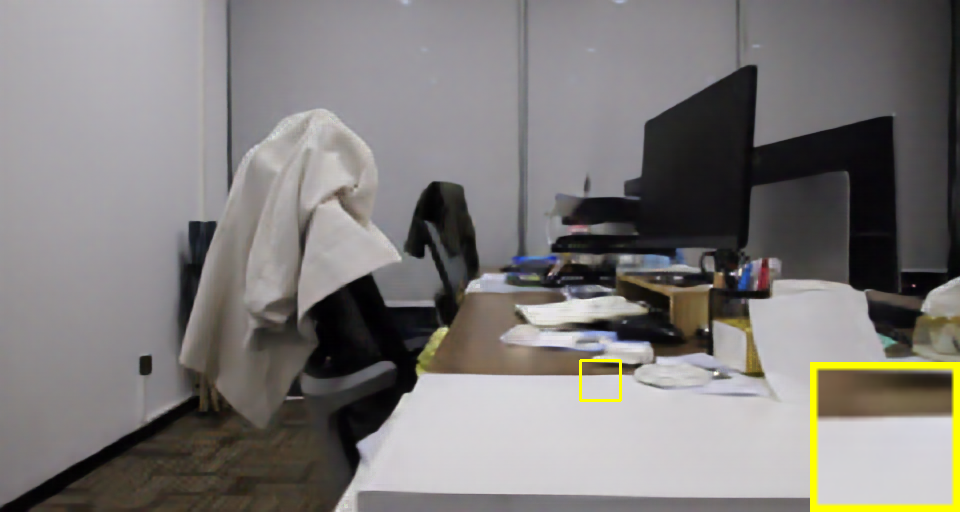}
		%\vspace{-1.5em}
		\caption{Ours {\scriptsize PSNR: \textbf{27.55}, SSIM: \textbf{0.89}}}
	\end{subfigure}
	\vspace{-0.15in}
	\caption{Qualitative comparison on the indoor videos in the SDSD dataset. 
		%Again, our result has sharper structures. 
		Please zoom to view.}
	\vspace{-0.15in}
	\label{fig:cmp_indoor}
\end{figure*}

\noindent\textbf{Baselines.} \
As far as we are aware of, there is no current work designed for directly mapping LNN videos to HNN videos.
So, we choose the following two classes of works to compare with.
First, we consider a rich collection of SOTA methods for video SR: BasicVSR~\cite{chan2021basicvsr}, IconVSR~\cite{chan2021basicvsr}, BasicVSR++~\cite{chan2021basicvsr++}, RBPN~\cite{haris2019recurrent}, Zooming~\cite{xiang2020zooming}, TGA~\cite{isobe2020video}, TDAN~\cite{tian2020tdan}, PFNL~\cite{yi2019progressive}, ToFlow~\cite{xue2019video}, and EDVR~\cite{wang2019edvr}.
We trained them on each dataset with their released code.
Second, we collectively use network models for video denoising, illumination enhancement, and SR in a cascaded manner:
illumination enhancement+SR, denoising+SR, illumination enhancement+denoise+SR, and denoise+illumination enhancement+SR, where ``+'' indicates the order of using different networks.
Here, we employ FastDVDnet~\cite{tassano2020fastdvdnet}, a SOTA method for video denoising, and TCE~\cite{zhang2021learning}, a SOTA method for video illumination enhancement, for use with various SOTA video SR methods.

%\newpage
%%\vspace*{2mm}
\noindent\textbf{Quantitative analysis.} \
Table~\ref{comparison1} shows the comparison results with the SOTA SR methods.
From the table, we can see that our method {\em consistently\/} achieves the highest PSNR and SSIM for all the datasets.
Especially, our PSNR values are higher than all others by a large margin.
This superiority shows that our method has strong capability of enhancing LLN videos.
Also, the right two columns show results on the SDSD indoor and outdoor subsets.
These videos contain dynamic scenes, so they are very challenging to handle.
Yet, our method is able to obtain high-quality results with top PSNR and SSIM for both subsets.

On the other hand, Table~\ref{comparison2} summarizes the comparison results with baselines that collectively combine SOTA video denoising, illumination enhancement, and SR networks.
Here, we trained each network (videos SR, illumination enhancement, and denoising) individually on the associated dataset.
From Table~\ref{comparison2}, we can see that our method always produces top PSNR values for all three datasets and our SSIM values stay high compared with others.

Fig.~\ref{comparison-speed} reports the run time of our method vs. the SOTA video SR methods.
We ran all methods on Intel 2.6GHz CPU \& TITAN XP GPU.
From the figure, we can see that our method is efficient with very low running time.

\begin{figure*}[t]
	\centering
	\begin{subfigure}[c]{0.24\textwidth}
		\centering
		\includegraphics[width=1.6in]{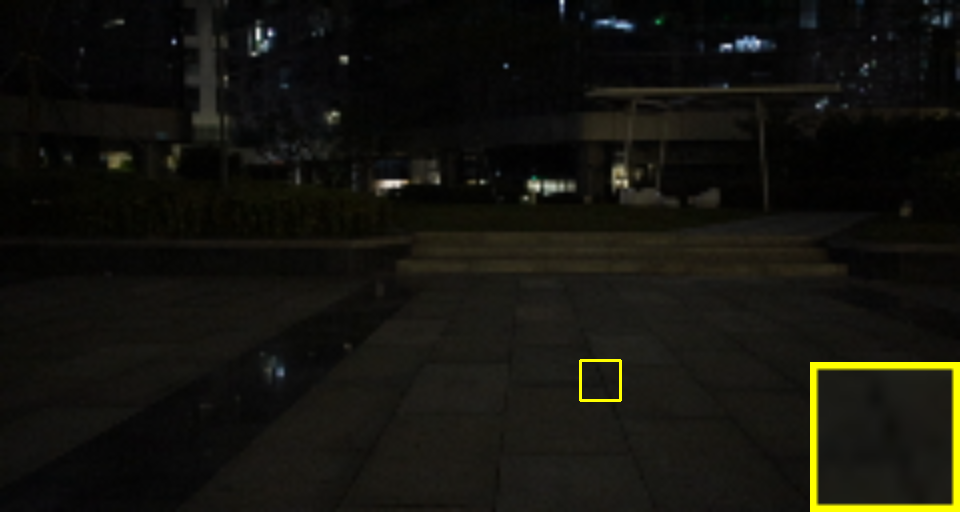}
		%\vspace{-1.5em}
		\caption{Input {\scriptsize PSNR: 11.30, SSIM: 0.65}}
	\end{subfigure}
	\begin{subfigure}[c]{0.24\textwidth}
		\centering
		\includegraphics[width=1.6in]{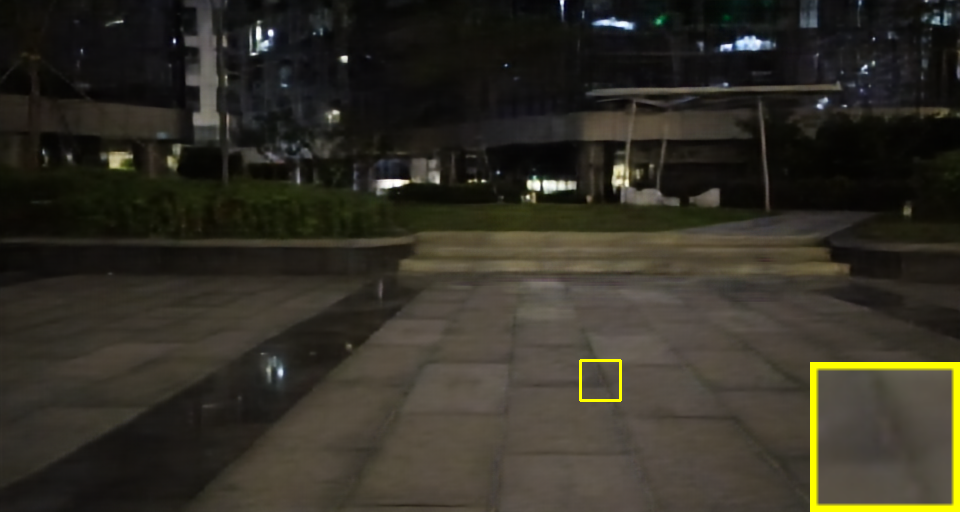}
		%\vspace{-1.5em}
		\caption{RBPN {\scriptsize PSNR: 24.59, SSIM: 0.83}}
	\end{subfigure}
	\begin{subfigure}[c]{0.24\textwidth}
		\centering
		\includegraphics[width=1.6in]{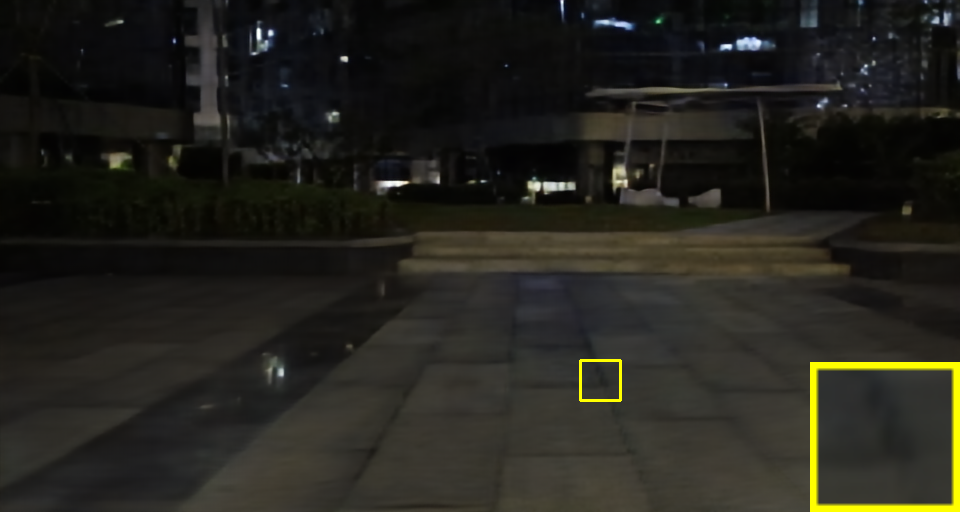}
		%\vspace{-1.5em}
		\caption{Zooming {\scriptsize PSNR: 24.24, SSIM: 0.82}}
	\end{subfigure}
	\begin{subfigure}[c]{0.24\textwidth}
		\centering
		\includegraphics[width=1.6in]{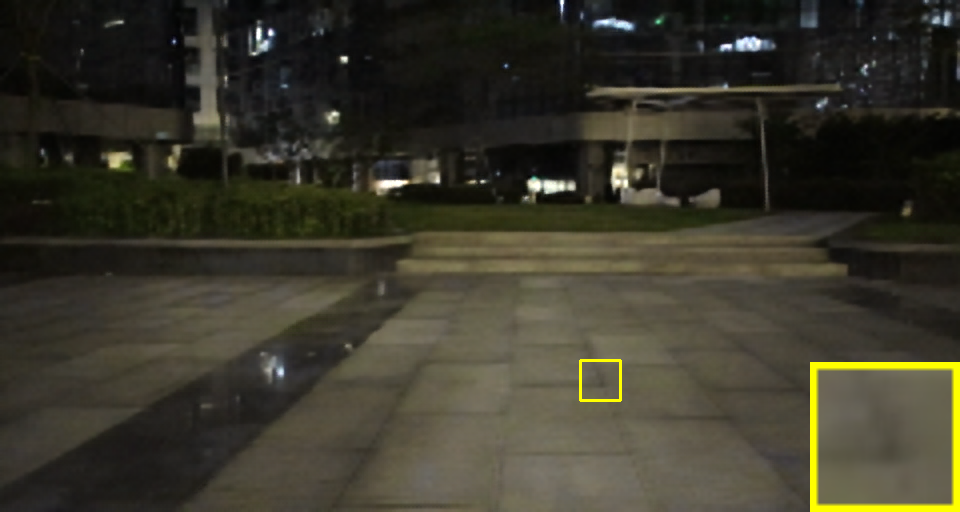}
		%\vspace{-1.5em}
		\caption{TGA {\scriptsize PSNR: 20.64, SSIM: 0.79}}
	\end{subfigure} 
	%\vspace{0.2em} 
	\\
	\begin{subfigure}[c]{0.24\textwidth}
		\centering
		\includegraphics[width=1.6in]{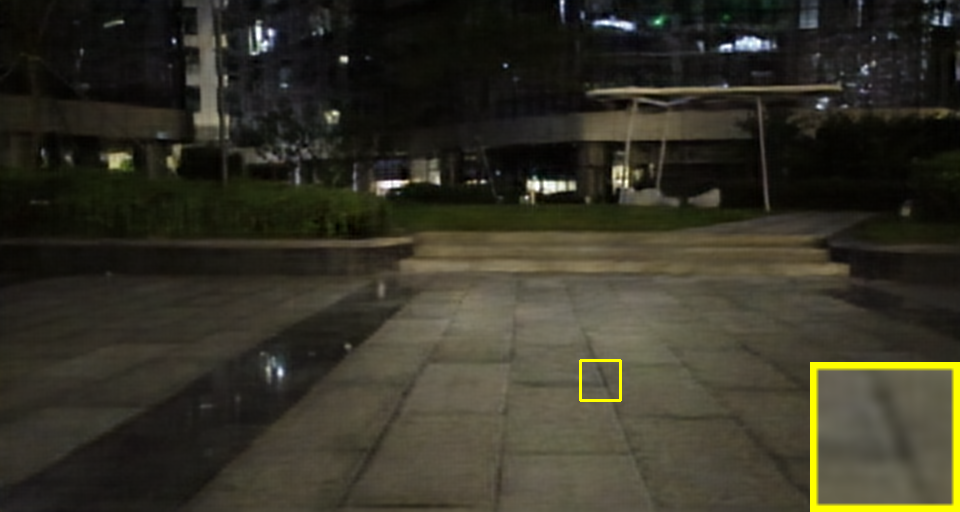}
		%\vspace{-1.5em}
		\caption{TDAN {\scriptsize PSNR: 22.37, SSIM: 0.81}}
	\end{subfigure} 
	\begin{subfigure}[c]{0.24\textwidth}
		\centering
		\includegraphics[width=1.6in]{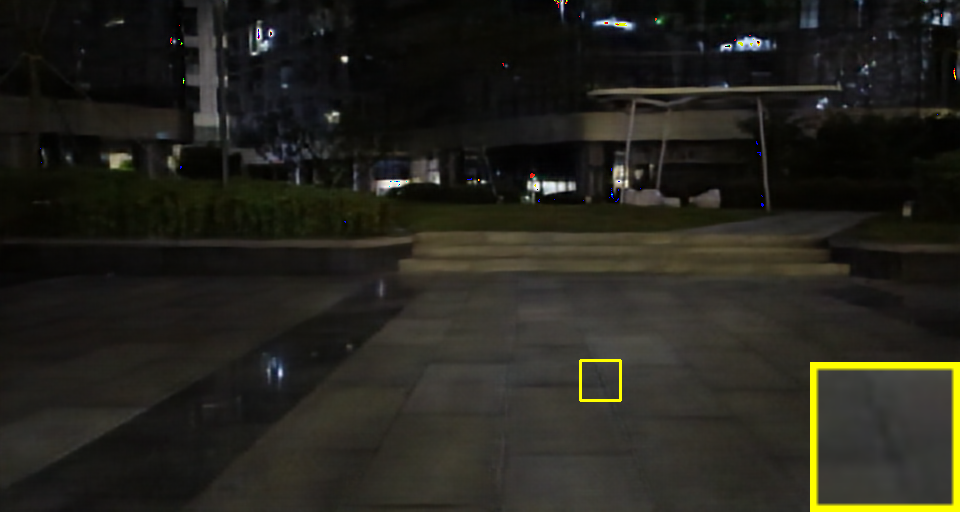}
		%\vspace{-1.5em}
		\caption{ToFlow {\scriptsize PSNR: 24.57, SSIM: 0.83}}
	\end{subfigure}
	\begin{subfigure}[c]{0.24\textwidth}
		\centering
		\includegraphics[width=1.6in]{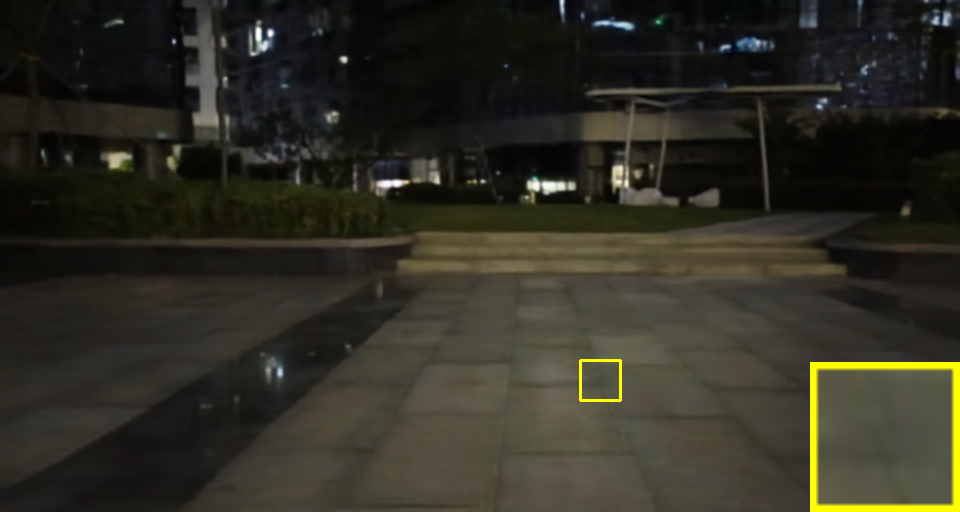}
		%\vspace{-1.5em}
		\caption{EDVR {\scriptsize PSNR: 24.64, SSIM: 0.83}}
	\end{subfigure}
	\begin{subfigure}[c]{0.24\textwidth}
		\centering
		\includegraphics[width=1.6in]{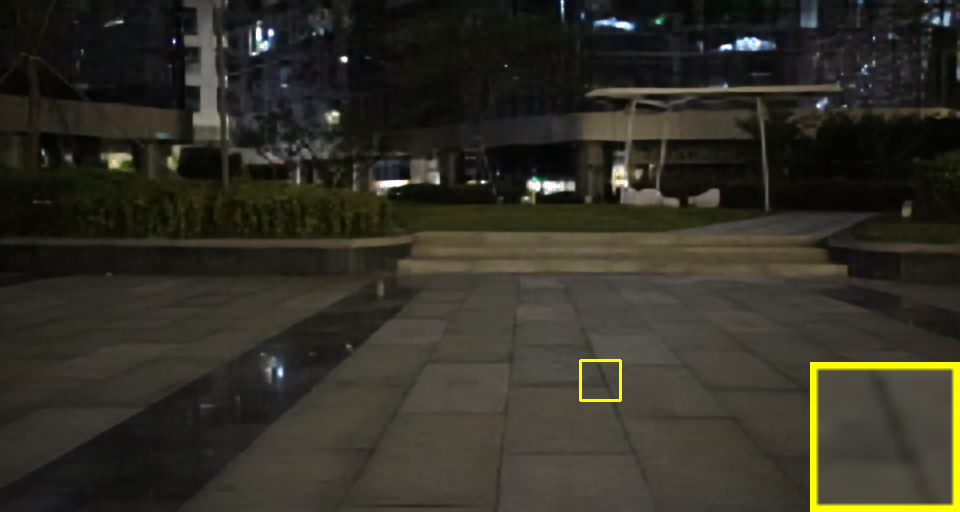}
		%\vspace{-1.5em}
		\caption{Ours {\scriptsize PSNR: \textbf{25.80}, SSIM: \textbf{0.84}}}
	\end{subfigure}
	\vspace{-0.15in}
	\caption{Qualitative comparison on the outdoor videos in the SDSD dataset. Please zoom to view the sample frames.}
	\label{fig:cmp_out}
	\vspace{-0.2in}
\end{figure*}

%%\vspace*{2mm}
\noindent\textbf{Qualitative analysis.} \
Next, we show visual comparisons with other methods.
Fig.~\ref{fig:cmp_smid} shows the comparison on SMID.
Overall, the results show two main advantages of our method over others.
First, the result from our method has high contrast and clear details, as well as natural color constancy and brightness.
Therefore, the frame processed by our method is more realistic than those by the others.
Second, in regions with complex textures, it can be observed that our outputs have fewer artifacts.
So, our result looks cleaner and sharper than those produced by the others.
Further, these results demonstrate that our method can {\em simultaneously} achieve video SR, noise reduction, and illumination enhancement.
%%; the corresponding effects are better than those of the baselines.

On the other hand, Figs.~\ref{fig:cmp_indoor} and~\ref{fig:cmp_out}, respectively, show the visual comparisons on the SDSD indoor and outdoor subsets that feature dynamic scenes.
Compared with the results of the baselines, our results are visually more appealing due to the explicit details, vivid colors, rational contrast, and plausible brightness.
These results show the limitations of the existing approaches in converting LLN videos to HNN videos, and the superiority of our framework.

%%\vspace*{2mm}
\noindent\textbf{User study.} \
Further, we conducted a large-scale user study with 80 participants (aged 18 to 52; 32 females and 48 males) to compare the perceptual quality of our method against various SOTA video SR approaches.
In detail, we randomly selected 36 videos from the test sets of SMID and SDSD, and compared the results of different methods on these videos using an AB test.
For each test video, our produced result is ``Video A'' whereas the result from some other baseline is ``Video B.'' 
In the test, each participant had to simultaneously watch videos A and B (we avoid bias by randomizing the left-right presentation order when showing videos A and B in each AB-test task) and choose among three options: ``I think Video A is better'', ``I think Video B is better'', and ``I cannot decide.''
Also, we asked the participants to make decisions based on the natural brightness, rich details, distinct contrast, and vivid color of the videos.
For each participant, the number of tasks is 10 methods $\times$ 2 videos $=20$, and it took around 30 minutes on average for each participant to complete the user study.

\begin{figure}[!t]
	\begin{center} 
		\includegraphics[width=1.0\linewidth]{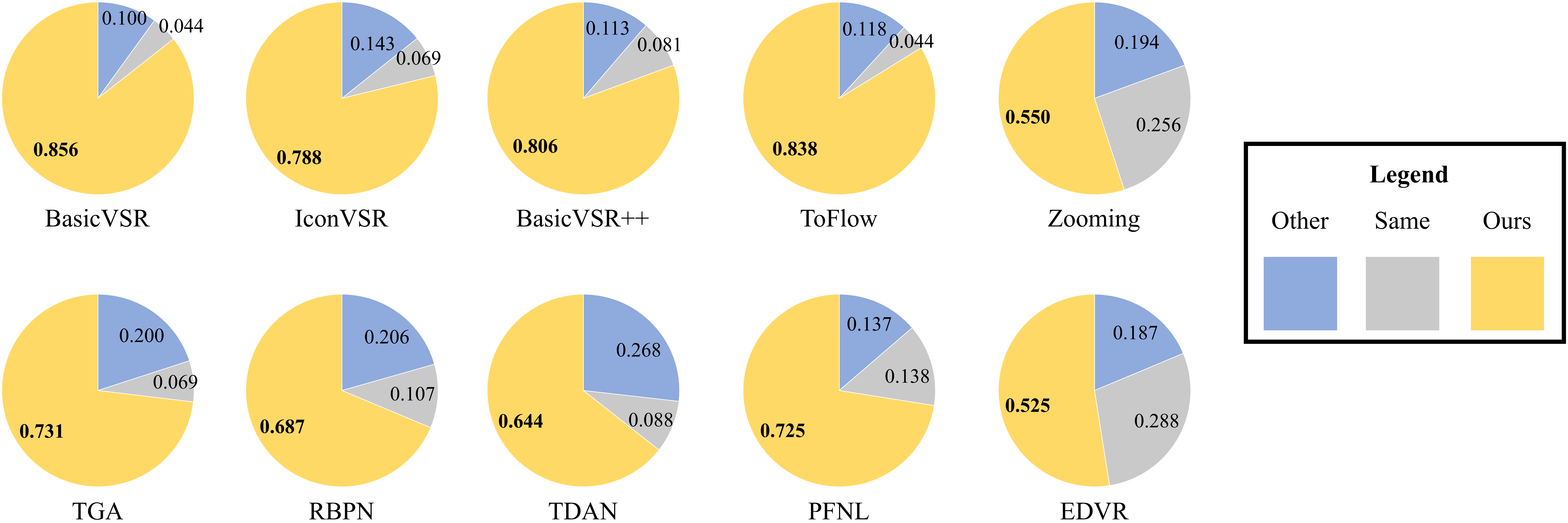}
	\end{center}
	\vspace{-0.15in}
	\caption{
		%The above pie charts summarize the results of our user study.
		``Ours'' is the percentage of test cases, in which the participant selected our results as better;
		``Other'' is the percentage that another method was chosen to be better; and
		``Same'' is the percentage that the user could not decide which one is better.
		%It is clear that the results from our method are more preferred by the participants
	}
	\label{us_tbl}
	\vspace{-0.25in}
\end{figure}

Fig.~\ref{us_tbl} summarizes the results of the user study, demonstrating that our results are more preferred by the participants over all the baselines.
Also, we performed the statistical analysis by using the T-TEST function in MS Excel and found that the associated p-values in the comparison with the baseline methods are all smaller than 0.001, showing that the conclusion has a significant level of 0.001 statistically.
%thereby showing that the conclusion has a significant level of 0.001 statistically.

\vspace{-0.1in}
\section{Conclusion}
This paper presents a new approach for video super resolution.
Our novel parametric representation, Deep Parametric 3D Filters (DP3DF), enables a direct mapping of LNN videos to HNN videos.
It intrinsically incorporates local spatiotemporal information and achieves video SR simultaneously with denoising and illumination enhancement efficiently within a single encoder-and-decoder network.
Besides, a dynamic residual frame can be jointly learned with the DP3DF, sharing the backbone and improving the visual quality of the results.

Extensive experiments were conducted on two real-world video datasets, SMID and SDSD, to show the effectiveness of our new approach.
Both the quantitative and qualitative comparisons between our approach and current SOTA methods demonstrate our approach's consistent top performance.
Further, an extensive user study with 80 participants was conducted to evaluate and compare the results in terms of human perception.
Results also showed that our results consistently receive higher ratings than those from the baselines.

\bibliography{aaai23}

\end{document}